\documentclass[11pt]{article}

\usepackage[margin=1in]{geometry}
\usepackage{amsmath,amssymb}
\usepackage{booktabs}
\usepackage{longtable}
\usepackage{graphicx}
\usepackage{xcolor}
\usepackage{colortbl}
\usepackage{hyperref}
\usepackage{url}
\usepackage[labelfont=bf]{caption}
\usepackage{float}

\newcommand{\noopsort}[1]{}

\hypersetup{
  colorlinks=true,
  linkcolor=blue!50!black,
  citecolor=blue!50!black,
  urlcolor=blue!50!black
}

\title{\textbf{Reconstruction}: A Blind Benchmark for Recovering Research Ideas from Pre-Publication Bibliographies\\
\large AI-Professor Project}

\author{
Shaolong Chen$^{1}$, Yanlin Fei$^{1}$, Nazhou Liu$^{1}$, Xinmiao Yu$^{1}$, Lei Li$^{1}$, Rahul Thapa$^{2}$,\\
Madalina Ciobanu$^{1}$, Navan Preet Singh$^{1}$, Qingqing Mao$^{1,3,*}$, Ritankar Das$^{1,3}$\\[0.6em]
\normalsize $^{1}$Prentis AI, San Francisco, CA \quad $^{2}$Stanford University, Stanford, CA\\
\normalsize $^{3}$Titan Holdings, San Francisco, CA\\\\
\normalsize $^{*}$Corresponding author: \texttt{qmao@prentis.ai}
}

\date{\today}

\begin{document}
\maketitle


\begin{abstract}
Can a language model recover the true research idea of a published paper when given only that paper's \emph{pre-publication} bibliography?
We introduce \textbf{Reconstruction}, a blind idea-recovery benchmark that withholds the seed paper and all contemporaneous or future literature, and asks models to propose hypotheses that an independent large language model judge matches against the held-out ground-truth idea.
A strict anti-leakage protocol---temporal citation cutoff, anonymous reference IDs, and frozen per-paper bibliographies---prevents \emph{prompt-time} leakage of the seed idea.
Across six scientific domains and 643 evaluated papers, seven frontier models achieve only modest Match rates of $3.4$--$15.0\%$.
We then evaluate a reference-only \textbf{multi-agent (top~4)} pipeline that combines cross-model review with a Swiss tournament over aligned hypothesis slots, without external web search.
Cross-model review plus tournament selection raises Match rates to $22.9$--$41.6\%$ across all six domains, with mean $36.0\%$ and an observed $2.4\times$ lift over the best dagger single-model baseline (top~4 proposers scored only by the other top~3 models).
The same qualitative pattern holds on papers first public after \mbox{2026-03-22}, the latest knowledge cutoff among the evaluated models, with Match rates $19.7$--$41.6\%$, mean $31.1\%$, and lift still $2.4\times$ vs.\ the same dagger baseline.
\end{abstract}

\section{Introduction}

Large language models (LLMs) are increasingly used to propose research ideas~\cite{lu2024aiscientist,yamada2025aiscientistv2,baek2025researchagent,si2024novelideas}.
Most evaluations ask whether a generated idea is novel, interesting, or preferred by humans.
A complementary question is harder and more diagnostic of literature understanding:
\emph{given only the references available before a paper was published, can a model recover what that paper actually proposed?}

We formalize this as \textbf{Reconstruction}.
For each seed paper, we build a blind reference context containing only literature published strictly before the seed's publication date $T_0$.
Models never see the seed title/abstract during generation; an LLM judge later compares model hypotheses to the held-out seed idea.
The primary metric is \textbf{Match rate}: the fraction of hypotheses that the judge labels as matching the ground truth.
Central to validity is the anti-leakage design (Section~\ref{sec:anti-leakage}): a hard temporal cutoff, anonymous reference IDs, information isolation from the seed, and frozen per-paper bibliographies shared by Default and multi-agent conditions.

\paragraph{Contributions.}
\begin{enumerate}
  \item \textbf{Reconstruction benchmark.} A time-cut, anti-leakage protocol for measuring idea recovery from pre-publication bibliographies across ML and five Nature-family domains.
  \item \textbf{Single-model baselines.} Seven frontier models on 643 papers; best average Match rate is $13.3\%\pm2.3\%$ (Claude-Opus-4.8), with domain scores typically in the $3.4$--$15.0\%$ band.
  \item \textbf{Reference-only multi-agent pipeline.} Cross-model review plus tournament selection raises Match rates to $22.9$--$41.6\%$ across all six domains, with mean $36.0\%$ and an observed $2.4\times$ lift over the best dagger single-model baseline (top~4 proposers scored only by the other top~3 models).
  We credit the full selection pipeline rather than collaboration alone; relative Default comparisons are reported as observed associations in the results, with candidate-matched controls left to follow-up work.
  \item \textbf{Contamination strata.} Date $\le$ knowledge cutoff is an upper bound on corpus reachability, not a demonstration of instance-level memorization (Section~\ref{sec:contam}, Appendix~\ref{app:contam}).
  The same qualitative pattern holds on the $n{=}236$ papers first public after \mbox{2026-03-22}, with Match rates $19.7$--$41.6\%$, mean $31.1\%$, and lift still $2.4\times$ vs.\ the same dagger baseline.
\end{enumerate}

\section{Related Work}

Automated scientific discovery and ideation systems use LLMs to propose research ideas from literature~\cite{lu2024aiscientist,yamada2025aiscientistv2,baek2025researchagent,zhao2026researchstudio,li2025chainofideas,zhao2025deepideation}.
Human studies ask whether LLM ideas are judged novel relative to expert proposals~\cite{si2024novelideas}, and follow-up work examines the gap between ideation and real execution outcomes~\cite{si2025ideationexecution}.
Benchmarks such as IdeaBench prompt models with the abstracts of a target paper's reference papers and rank generated ideas against the target's actual idea~\cite{guo2024ideabench}, while RINoBench targets automated novelty judgment of research ideas~\cite{schopf2026rinobench}. A related rediscovery line asks models to regenerate a known paper's hypothesis from curated, paper-derived background and inspiration material: MOOSE-Chem in chemistry~\cite{yang2025moosechem}, ResearchBench across twelve disciplines~\cite{liu2025researchbench}, and AI Idea Bench 2025 on AI papers~\cite{qiu2025aiideabench}. Closest in conditioning, Chen et al.\ reverse-engineer inspirational prior work for published papers and measure how far LLM-generated ideas remain from human ideas~\cite{chen2026measuringgap}.
Our setting differs: we measure \emph{recovery} of a known published idea under a strict temporal information cutoff, rather than open-ended idea generation or novelty ranking. Relative to the rediscovery benchmarks above, proposers here see \emph{only} the seed's full pre-publication bibliography (which is frozen, anonymized, and temporally cut, with no seed-derived background or curated inspiration set) and recovery is scored as a blind binary Match rate.

Temporal evaluation is closely related.
HindSight restricts ideation to pre-cutoff literature and scores ideas against post-cutoff future publications~\cite{jiang2026hindsight}; Reconstruction instead recovers the seed paper's own idea from its pre-publication bibliography alone.
Parametric memorization of that seed remains a separate concern~\cite{morris2025memorize}.

Multi-agent debate can improve factuality and reasoning~\cite{du2023debate}.
We adapt a related idea---cross-model review plus Swiss-system selection~\cite{csato2017swiss}---to Reconstruction. Because multi-agent selects five hypotheses from a larger candidate pool, its gains also relate to inference-time scaling: repeated sampling with
selection~\cite{brown2024monkeys,snell2024testtime}, self-consistency~\cite{wang2023selfconsistency}, and mixture-of-agents aggregation~\cite{wang2024moa} improve accuracy by generating and choosing among more candidates. Our design addresses this axis explicitly: Table~\ref{tab:candbound} bounds the contribution of candidate count on the frozen Default grid, and the Default vs.\ multi-agent comparison is reported as an observed association. Hypothesis generation in both the Default and multi-agent conditions uses only the same frozen blind bibliography $\mathcal{R}_{<T_0}$ and makes no runtime web-search calls.
Final Match scoring is a separate evaluation step for both conditions: an independent LLM judge compares each hypothesis with the held-out seed title and abstract, with leave-one-out/origin recusal to avoid self-evaluation bias.

\section{The Reconstruction Benchmark}

\subsection{Task}
Given a seed paper $s$ with publication date $T_0$, construct a blind corpus
\[
\mathcal{R}_{<T_0} = \{\, r : \mathrm{published}(r) < T_0 \,\}
\]
from $s$'s bibliography (resolved to title/abstract; anonymous IDs \texttt{ref-001}, \ldots).
The seed itself and any same-day/future literature are held out.
A reconstruction proposer produces $n_s{=}5$ distinct hypotheses $\{h_i\}_{i=1}^{n_s}$, each with supporting reference IDs.
Papers that fail to yield five hypotheses under multi-agent hypothesis generation for every top~4 model are excluded from the reported seed-paper set (Table~\ref{tab:funnel}), so every scored case has $n_s{=}5$.
A judge $J$ that sees only the seed title/abstract and hypothesis title/summary returns a binary match label $m_i(s;P,J)\in\{0,1\}$.
For proposer $P$ on seed paper $s$ under judge $J$, the paper-level Match rate is
\[
\mathrm{Match}(s;P,J)=\frac{1}{n_s}\sum_{i=1}^{n_s} m_i(s;P,J).
\]
Let $\mathcal{S}$ be the seed-paper set and $\{\mathcal{S}_d\}_{d\in\mathcal{D}}$ its partition into domains (here $|\mathcal{D}|{=}6$: ML, Astronomy, Chemistry, Materials, Medicine, Physics), so $\mathcal{S}=\bigcup_{d\in\mathcal{D}}\mathcal{S}_d$.
The domain-level score of $P$ under $J$ is
\[
\mathrm{Match}_d(P,J)=\frac{1}{|\mathcal{S}_d|}\sum_{s\in\mathcal{S}_d}\mathrm{Match}(s;P,J),
\]
and the overall score on the full seed-paper set is
\[
\mathrm{Match}(P,J)=\frac{1}{|\mathcal{S}|}\sum_{s\in\mathcal{S}}\mathrm{Match}(s;P,J).
\]
For a Default proposer $P$, let $\mathcal{J}(P)$ be its judge panel: the six leave-one-out models for primary rows or the other three top models for dagger top~4 rows.
The reported domain score averages the already-defined per-judge domain scores over that panel:
\begin{align*}
\mathrm{Match}_d(P)
&=\frac{1}{|\mathcal{J}(P)|}\sum_{J\in\mathcal{J}(P)}\mathrm{Match}_d(P,J),\\
\sigma_d(P)
&=\sqrt{\frac{1}{|\mathcal{J}(P)|}\sum_{J\in\mathcal{J}(P)}\Bigl(\mathrm{Match}_d(P,J)-\mathrm{Match}_d(P)\Bigr)^2}.
\end{align*}
Each Default domain cell in Table~\ref{tab:main} reports $\mathrm{Match}_d(P)\pm\sigma_d(P)$.

Multi-agent requires hypothesis-level origin recusal, and the proposer of each champion is not fixed in advance: Swiss selection chooses which model wins each slot, so the eligible judge set varies with the champion.
Every reported multi-agent case has exactly $n_s{=}5$ champions (Table~\ref{tab:funnel}).
Let $o(s,i)$ be the origin of champion $i$, and let $\mathcal{J}_{\mathrm{top4}}$ be the top~4 judge panel.
A recused judgment on an origin-matched champion is treated as missing, not as a No-match label.
For each judge $J$, restrict to champions $J$ did not originate and average first over those champions (within each seed) and then over seeds, parallel to $\mathrm{Match}_d(P,J)$:
\begin{align*}
\mathrm{Match}(s;\mathrm{MA},J)
&=\frac{1}{|\{i:o(s,i)\neq J\}|}
\sum_{i:\,o(s,i)\neq J}m_i(s;\mathrm{MA},J),\\
\mathrm{Match}_d(\mathrm{MA},J)
&=\frac{1}{|\mathcal{S}_d|}\sum_{s\in\mathcal{S}_d}\mathrm{Match}(s;\mathrm{MA},J).
\end{align*}
(If a seed has no champion eligible for $J$, that seed is omitted from $\mathrm{Match}_d(\mathrm{MA},J)$.)
The four scores $\{\mathrm{Match}_d(\mathrm{MA},J)\}_{J\in\mathcal{J}_{\mathrm{top4}}}$ are then aggregated as a weighted set, with weight equal to the number of eligible (seed, champion) pairs for that judge:
\[
w_J
=\sum_{s\in\mathcal{S}_d}|\{i:o(s,i)\neq J\}|,
\qquad
\mathrm{Match}_d(\mathrm{MA})
=\frac{\sum_{J\in\mathcal{J}_{\mathrm{top4}}}w_J\,\mathrm{Match}_d(\mathrm{MA},J)}{\sum_{J\in\mathcal{J}_{\mathrm{top4}}}w_J},
\]
\[
\sigma_d(\mathrm{MA})
=\sqrt{
\frac{\sum_{J\in\mathcal{J}_{\mathrm{top4}}}w_J\bigl(\mathrm{Match}_d(\mathrm{MA},J)-\mathrm{Match}_d(\mathrm{MA})\bigr)^2}
{\sum_{J\in\mathcal{J}_{\mathrm{top4}}}w_J}
}.
\]
Thus $\mathrm{Match}_d(\mathrm{MA})$ uses the same outer order as $\mathrm{Match}_d(P)$---champions, then seeds, then judges---except that the final judge average is weighted by eligible coverage, because judges who originate fewer champions evaluate a larger eligible set.
The multi-agent domain cell reports $\mathrm{Match}_d(\mathrm{MA})\pm\sigma_d(\mathrm{MA})$.
Empirically $\sigma_d(\mathrm{MA})$ is much larger than typical $\sigma_d(P)$ (Table~\ref{tab:main}).
This is expected under the definition of $\sigma_d$ as dispersion \emph{across judges}, not across papers.
Multi-agent concentrates stronger, more specific champions, so the top~4 judges disagree substantially on those champions (e.g., Claude/GPT are systematically more permissive than Kimi/GLM).
Default single-model hypotheses are weaker and more often unanimously unmatched, which compresses judge-to-judge dispersion and keeps $\sigma_d(P)$ small.

For any reported row $Q$, let $M_d(Q)$ denote its domain point estimate---$\mathrm{Match}_d(P)$ for Default or $\mathrm{Match}_d(\mathrm{MA})$ for multi-agent.
The Avg column is
\[
\mathrm{Avg}(Q)
=\frac{1}{|\mathcal{D}|}\sum_{d\in\mathcal{D}}M_d(Q),
\qquad
\sigma_{\mathrm{Avg}}(Q)
=\sqrt{\frac{1}{|\mathcal{D}|}\sum_{d\in\mathcal{D}}\Bigl(M_d(Q)-\mathrm{Avg}(Q)\Bigr)^2},
\]
shown as $\mathrm{Avg}(Q)\pm\sigma_{\mathrm{Avg}}(Q)$.
Both $\sigma_d$ and $\sigma_{\mathrm{Avg}}$ divide by the full count rather than applying a Bessel correction, since the judge panel and the six domains are the complete sets being described rather than samples from a larger population.
This domain-unweighted Avg is not the same as pooling over $\mathcal{S}$, because domains have unequal $|\mathcal{S}_d|$.

\subsection{Anti-leakage design}
\label{sec:anti-leakage}
Reconstruction is designed so that recovering the seed idea requires reasoning over the pre-publication bibliography rather than reading the seed or contemporaneous literature \emph{at prompt time}.
The protocol does not prevent a seed from having entered pretraining (Section~\ref{sec:contam}).
\begin{itemize}
  \item \textbf{Temporal cutoff:} only $\mathrm{published}<T_0$ enters $\mathcal{R}_{<T_0}$; undated references are excluded.
  \item \textbf{Information isolation:} proposers never see the seed paper or contemporaneous literature.
  \item \textbf{Anonymous references:} bibliography entries are exposed only as opaque IDs (\texttt{ref-001}, \ldots) with title/abstract text---no venue shortcuts that would trivially reveal the seed.
  \item \textbf{Frozen per-paper bibliographies:} the multi-agent pipeline reuses the exact $\mathcal{R}_{<T_0}$ resolved in Default, so the two conditions cannot diverge by resolving a different reading list.
  \item \textbf{Evidence binding:} each hypothesis must cite anonymous supporting references.
  \item \textbf{Judge self-evaluation avoidance:} the judge model must differ from the proposer; under multi-agent, judges recuse on hypotheses they originated.
  \item \textbf{Prompt-time vs.\ parametric:} the items above prevent prompt-time leakage; they do not bound what a model may already store in its weights. Section~\ref{sec:contam} and Appendix~\ref{app:contam} stratify Match rates by earliest public date versus each model's knowledge cutoff.
\end{itemize}

\subsection{Default (single-model) protocol}
In a single generation call, each model is asked to produce five distinct hypotheses from the same blind references and is scored as its own case.
No tournament is used in Default mode.

\subsection{Multi-agent (top~4) protocol}
We select the four strongest Default models by six-domain average Match rate on the reported seed-paper set: Claude-Opus-4.8, GPT-5.6-Sol-Pro, Kimi-K3, and GLM-5.2.
This is a \emph{post-selected} ensemble relative to the same papers used for scoring; we therefore interpret multi-agent results as evaluating this fixed top~4 roster rather than a selection rule validated on a held-out split.
\begin{enumerate}
  \item \textbf{Parallel generation:} each model, in one generation call, produces five distinct hypotheses from identical frozen blind references.
  \item \textbf{Slot alignment:} for slot $k\in\{1,\ldots,5\}$, collect the $k$-th hypothesis from each model (4 candidates per slot; 20 candidates total before selection).
  \item \textbf{Reference-only review:} other models review each candidate (proposer recusal), and the proposing model may refine its candidate in response to that review; no web search.
  \item \textbf{Swiss selection:} candidates in each slot enter a Swiss tournament judged only on blind references, with conflict-of-interest recusal and presentation-order debiasing (Appendix~\ref{app:swiss}).
  \item \textbf{Assemble:} five slot champions form one \texttt{multi-agent} case, scored by the final Match judge independently of prior review/tournament calls. For each champion, the origin model recuses; its missing label is excluded rather than counted as No-match (Appendix~\ref{app:judge}).
\end{enumerate}
The multi-agent pipeline freezes per-paper $\mathcal{R}_{<T_0}$ from completed Default runs so it cannot resolve a different bibliography.

\section{Experiments}

\subsection{Dataset construction and filtering}
\label{sec:dataset}

\paragraph{Seed collection.}
We collect seed papers as title lists from six sources (879 titles in total).
The ML set is the full ICML 2026 Oral program ($N{=}168$), scraped from the conference website.
The remaining five domains are scraped from the corresponding Nature-family journal pages on 14~July~2026:
\emph{Nature Astronomy} (91),
\emph{Nature Chemistry} (115),
\emph{Nature Materials} (131),
\emph{Nature Medicine} (221),
and \emph{Nature Physics} (153).
Each CSV provides a title (Nature lists also include a citation-count field used only for provenance, not for filtering).

\paragraph{Metadata and bibliography resolution.}
Each title is resolved to bibliographic metadata (title, abstract, authors, publication date, DOI when available) via Semantic Scholar as the primary lookup, with fallbacks to OpenReview (especially for ICML), Crossref, OpenAlex, and arXiv.
We then assemble the seed's bibliography by merging reference strings from Semantic Scholar, Crossref, and OpenAlex, and---when a PDF is obtainable---from arXiv or OpenReview PDF extraction.
References are normalized, deduplicated, resolved to title/abstract/date, and assigned anonymous IDs (\texttt{ref-001}, \ldots).
The temporal cutoff is $T_0{=}{}$the seed's publication date: only references with $\mathrm{published}(r)<T_0$ enter the blind reconstruction context $\mathcal{R}_{<T_0}$; undated references and the seed itself are held out.

\paragraph{Eligibility criteria.}
A seed is retained only if (i)~it resolves to a real paper with a usable publication date, and (ii)~at least three references are both successfully resolved and published strictly before $T_0$ (minimum bibliography size for hypothesis generation).
Seeds that fail these checks are excluded before or during the Default run and do not contribute to Match-rate statistics.

\paragraph{Retention funnel.}
Table~\ref{tab:funnel} summarizes how raw CSV titles become the evaluated set, including reasons for launched-but-incomplete seeds.
Of 879 collected titles, 53 never enter a Default run (eligibility failures or composer deselection before launch), so $879{-}53{=}826$ seeds are launched.
Among launched seeds, Default completes 745 papers, leaving $826{-}745{=}81$ incomplete runs (Table~\ref{tab:funnel}, Incomplete-by-reason columns).
Medicine accounts for 35 of these: the Default job launched 195 seeds but stopped early at a target of 160 completed papers (\texttt{early\_stop\_target}${}={}$160), primarily to save wall-clock time and API cost, leaving papers 161--195 unfinished.
Separately, within some \emph{completed} Medicine papers, Claude-Opus-4.8 occasionally returned empty outputs with OpenRouter \texttt{finish\_reason=content\_filter} (4 papers in the Medicine audit); those cases mark only that proposer as unavailable and do \emph{not} remove the paper from Default OK.

\paragraph{Reported $n$ and multi-agent alignment.}
Multi-agent runs \emph{derive} from completed Default runs and freeze the same per-paper $\mathcal{R}_{<T_0}$.
Materials loses one additional paper under multi-agent to a provider \texttt{content\_filter} refusal, yielding 120 multi-agent completions before the $n_s{=}5$ filter.
For Medicine, Default completes 160 papers, but multi-agent evaluates only the first 80 to control API cost.
Across domains, 21 further papers are excluded because multi-agent hypothesis generation returned fewer than five hypotheses for at least one top~4 model, so the Swiss slot count $\min_m|\mathrm{hyps}_m|$ was strictly less than five (Table~\ref{tab:funnel}, MA~$n_s{<}5$ column).
The reported seed-paper set therefore retains only papers with exactly $n_s{=}5$ champions, with size $n{=}643$.

\begin{table*}[t]
\centering
\scriptsize
\caption{Seed retention from CSV titles to the reported seed-paper set, with launched-but-incomplete failures broken out by reason.
\emph{CSV}: parsed titles;
\emph{Not launched}: excluded before Default \texttt{run\_started};
\emph{Incomplete (by reason)}: launched but not completed; the four columns sum to $81{=}826{-}745$ (CSV $-$ Not launched $-$ Default OK);
\emph{Unresolved}: title unresolved; \emph{References$<3$}: fewer than three pre-$T_0$ references; \emph{No date}: no usable publication date; \emph{Early stop}: launched papers left unfinished when the job hit an early-stop completion target chosen to limit time and cost (all 35 are Medicine: 195 launched, stop at 160);
\emph{Default OK}: completed blind-bibliography reconstruction;
\emph{MA $n_s{<}5$}: multi-agent--aligned papers dropped because at least one top~4 hypothesis-generation call returned fewer than five hypotheses (Swiss slot count $=\min_m|\mathrm{hyps}_m|{<}5$);
\emph{Reported $n$}: counts used in Table~\ref{tab:main} (multi-agent--aligned with $n_s{=}5$).
$^{a}$Materials: 121 Default OK, minus one multi-agent \texttt{content\_filter} refusal and three MA~$n_s{<}5$ drops $\Rightarrow$ Reported $n{=}117$.
$^{b}$Medicine: Default completes 160; multi-agent evaluates the first 80 for cost, then drops two MA~$n_s{<}5$ papers $\Rightarrow$ Reported $n{=}78$.
\textbf{Warning:} Claude-Opus-4.8 can return empty hypotheses under OpenRouter \texttt{content\_filter}; in the Medicine Default audit this affected 4 completed papers (proposer marked unavailable; other models still scored). This is distinct from the Early-stop and MA~$n_s{<}5$ columns.}
\label{tab:funnel}
\begin{tabular}{@{}lrrrrrrrrr@{}}
\toprule
& & & \multicolumn{4}{c}{Incomplete (by reason)} & & & \\
\cmidrule(lr){4-7}
Domain & CSV & Not launched & Unresolved & References$<3$ & No date & Early stop & Default OK & MA $n_s{<}5$ & Reported $n$ \\
\midrule
ML & 168 & 0 & 24 & 18 & 1 & 0 & 125 & 5 & 120 \\
Astronomy & 91 & 4 & 0 & 0 & 0 & 0 & 87 & 2 & 85 \\
Chemistry & 115 & 4 & 1 & 1 & 0 & 0 & 109 & 4 & 105 \\
Materials & 131 & 9 & 0 & 1 & 0 & 0 & 121 & 3 & $117^{a}$ \\
Medicine & 221 & 26 & 0 & 0 & 0 & 35 & 160 & 2 & $78^{b}$ \\
Physics & 153 & 10 & 0 & 0 & 0 & 0 & 143 & 5 & 138 \\
\midrule
Total & 879 & 53 & 25 & 20 & 1 & 35 & 745 & 21 & 643 \\
\bottomrule
\end{tabular}
\end{table*}

\paragraph{Infrastructure.}
LLM calls use OpenRouter.
Literature APIs: Semantic Scholar (authenticated), OpenAlex and Crossref (polite-pool contact emails), OpenReview (enabled; no session cookie), and arXiv PDF download when an identifier is available.
Reference resolution runs with concurrency~8; reconstruction judges with concurrency~5.
Default Nature-domain runs started on 15~July~2026; multi-agent derives followed on 22--27~July~2026.

\subsection{Models and judging}
We evaluate seven OpenRouter model snapshots (evaluation window July~2026), listed with provider, display name, release date, and provider ID:
\begin{itemize}
  \item Anthropic: Claude Opus~4.8 (2026-05-27); \texttt{anthropic/claude-opus-4.8}
  \item OpenAI: GPT-5.6 Sol Pro (2026-07-09); \texttt{openai/gpt-5.6-sol-pro}
  \item MoonshotAI: Kimi~K3 (2026-07-16); \texttt{moonshotai/kimi-k3}
  \item Z.ai: GLM~5.2 (2026-06-16); \texttt{z-ai/glm-5.2}
  \item Google: Gemini~3.1 Pro Preview (2026-02-19); \texttt{google/gemini-3.1-pro-preview}
  \item DeepSeek: DeepSeek-V4-Pro (2026-04-24); \texttt{deepseek/deepseek-v4-pro}
  \item Qwen: Qwen3.7-Max (2026-05-21); \texttt{qwen/qwen3.7-max}
\end{itemize}
IDs are written as resolved at run time; provider aliases may evolve, so we treat the July~2026 OpenRouter routing as the snapshot of record.
Release dates follow the project's model-release registry (and UI metadata) used for temporal cutoff checks.
For Default, each proposer's Match rate is averaged over the other six models as judges (cell mean$\pm$std across those judges).
For multi-agent, the four top models serve as judges with origin-based recusal.
To make Default top~4 scores more comparable to that panel, Table~\ref{tab:main} also reports four extra rows in which each top~4 proposer is scored only by the other three top models.
The final Match judge prompt and binary rubric are given in Appendix~\ref{app:judge}, with one borderline Match example and one No-match example in Table~\ref{tab:borderline}.

\subsection{Main results}
Table~\ref{tab:main} summarizes Match rates.
Single models remain in a low band: observed cell means in the primary rows span $3.4$--$15.0\%$, and the best six-domain average is $13.3\%\pm2.3\%$ (Claude-Opus-4.8; Avg $\pm$ is across domains).
Restricting Default top~4 proposers to the other-top~3 judge panel slightly changes point estimates (typically within a few points) but does not move them into the multi-agent band.
The multi-agent (top~4) pipeline---cross-model review plus Swiss selection---reaches $22.9\%$ (ML), $36.5\%$ (Astronomy), $38.4\%$ (Chemistry), $40.1\%$ (Materials), $41.6\%$ (Medicine), and $36.4\%$ (Physics), averaging $36.0\%\pm6.1\%$ across domains.

\begin{table*}[t]
\centering
\scriptsize
\caption{Reconstruction Match rates (\%).
For Default rows, cells show mean$\pm$std across leave-one-out judges (six judges, or three peer top models in the dagger block).
For multi-agent, the center is the weight\-ed judge-set aggregate $\mathrm{Match}_d(\mathrm{MA})$; $\pm$ is the corresponding weighted $\sigma_d(\mathrm{MA})$.
The Avg column reports the unweighted mean across the six domain columns, with $\pm$ the standard deviation \emph{across domains} (not across judges).
Within each Default block (all seven models; dagger top~4), bold marks the column maximum.
\texttt{vs best single} is multi-agent divided by the best dagger (top~4, other-top~3 judges) single-model score in that column.}
\label{tab:main}
\begin{tabular}{@{}lccccccc@{}}
\toprule
Model & ML$_{120}$ & Astro$_{85}$ & Chem$_{105}$ & Mat$_{117}$ & Med$_{78}$ & Phys$_{138}$ & Avg \\
\midrule
Claude-Opus-4.8
  & $\mathbf{8.2}{\pm}2.4$ & $\mathbf{14.7}{\pm}3.1$ & $14.0{\pm}5.3$ & $\mathbf{14.0}{\pm}3.7$ & $14.2{\pm}3.0$ & $\mathbf{15.0}{\pm}3.2$ & $\mathbf{13.3}{\pm}2.3$ \\
GPT-5.6-Sol-Pro
  & $7.6{\pm}1.3$ & $12.9{\pm}2.6$ & $\mathbf{15.0}{\pm}3.8$ & $13.8{\pm}2.9$ & $\mathbf{14.6}{\pm}2.7$ & $12.7{\pm}2.1$ & $12.8{\pm}2.4$ \\
Kimi-K3
  & $7.9{\pm}1.9$ & $10.7{\pm}2.6$ & $9.8{\pm}2.6$ & $10.1{\pm}2.2$ & $11.1{\pm}1.9$ & $10.7{\pm}2.4$ & $10.0{\pm}1.0$ \\
GLM-5.2
  & $7.2{\pm}1.2$ & $9.5{\pm}2.8$ & $9.8{\pm}3.0$ & $9.2{\pm}2.3$ & $10.1{\pm}2.0$ & $10.2{\pm}2.5$ & $9.3{\pm}1.0$ \\
Gemini 3.1-Pro-Preview
  & $5.9{\pm}0.6$ & $8.9{\pm}1.6$ & $10.2{\pm}2.6$ & $8.4{\pm}2.2$ & $10.5{\pm}1.8$ & $9.2{\pm}1.7$ & $8.9{\pm}1.5$ \\
DeepSeek-V4-Pro
  & $3.4{\pm}1.3$ & $6.7{\pm}2.5$ & $6.6{\pm}2.8$ & $7.2{\pm}3.0$ & $6.9{\pm}2.2$ & $7.0{\pm}2.6$ & $6.3{\pm}1.3$ \\
Qwen3.7-Max
  & $4.6{\pm}1.3$ & $5.3{\pm}2.2$ & $8.4{\pm}2.5$ & $5.9{\pm}2.4$ & $4.8{\pm}1.8$ & $6.7{\pm}1.8$ & $5.9{\pm}1.3$ \\
\midrule
\multicolumn{8}{@{}l@{}}{\textit{Top~4 proposers, judged only by the other top~3 models}} \\
Claude-Opus-4.8$^{\dagger}$
  & $8.8{\pm}3.0$ & $\mathbf{15.5}{\pm}4.0$ & $16.3{\pm}6.0$ & $\mathbf{15.4}{\pm}4.6$ & $15.3{\pm}3.3$ & $\mathbf{15.8}{\pm}3.8$ & $\mathbf{14.5}{\pm}2.6$ \\
GPT-5.6-Sol-Pro$^{\dagger}$
  & $7.9{\pm}1.6$ & $13.6{\pm}3.3$ & $\mathbf{16.8}{\pm}4.3$ & $15.2{\pm}2.8$ & $\mathbf{15.8}{\pm}2.7$ & $13.9{\pm}2.2$ & $13.9{\pm}2.9$ \\
Kimi-K3$^{\dagger}$
  & $\mathbf{9.3}{\pm}1.6$ & $11.1{\pm}3.0$ & $11.4{\pm}2.9$ & $11.3{\pm}1.6$ & $12.1{\pm}2.2$ & $11.4{\pm}2.8$ & $11.1{\pm}0.9$ \\
GLM-5.2$^{\dagger}$
  & $8.1{\pm}0.3$ & $11.7{\pm}2.1$ & $12.3{\pm}2.2$ & $11.2{\pm}1.4$ & $11.9{\pm}0.9$ & $12.2{\pm}1.9$ & $11.2{\pm}1.5$ \\
\midrule
\rowcolor{magenta!8}
multi-agent (top 4)
  & $\mathbf{22.9}{\pm}6.0$ & $\mathbf{36.5}{\pm}11.4$ & $\mathbf{38.4}{\pm}8.5$ & $\mathbf{40.1}{\pm}10.3$ & $\mathbf{41.6}{\pm}10.3$ & $\mathbf{36.4}{\pm}10.4$ & $\mathbf{36.0}{\pm}6.1$ \\
\rowcolor{magenta!5}
vs best single ($\times$)
  & $2.5\times$ & $2.4\times$ & $2.3\times$ & $2.6\times$ & $2.6\times$ & $2.3\times$ & $2.4\times$ \\
\bottomrule
\end{tabular}
\end{table*}

\begin{figure}[t]
\centering
\includegraphics[width=\linewidth]{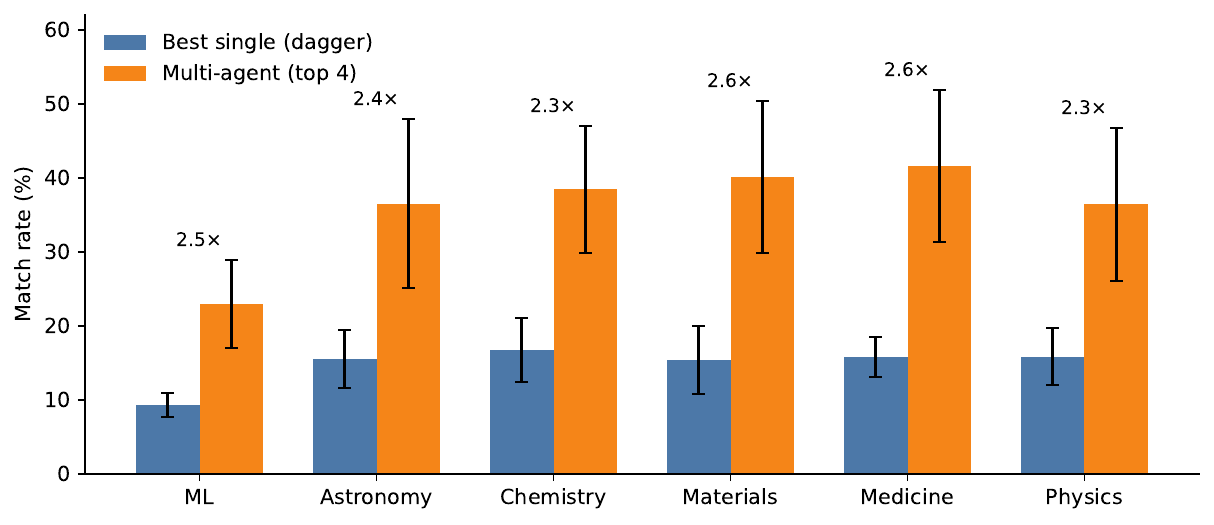}
\caption{Best top~4 single-model Match rate (dagger panel) vs.\ multi-agent (top~4) on six domains.
Both bars include error bars showing the population standard deviation across their respective eligible judge scores.
Annotations show the lift ratio.}
\label{fig:lift}
\end{figure}

Figure~\ref{fig:lift} visualizes the lift relative to the best dagger (top~4, other-top~3 judges) single-model score per domain.
Relative to that panel, the multi-agent pipeline's observed ratios are $2.5\times$ (ML), $2.4\times$ (Astronomy), $2.3\times$ (Chemistry), $2.6\times$ (Materials), $2.6\times$ (Medicine), and $2.3\times$ (Physics); the mean of these ratios is $2.4\times$.
Materials and Medicine sit at the high end of the multi-agent absolute range.
We treat these ratios as an \emph{observed association for the full pipeline}, not an isolated collaboration effect:
multi-agent selects 5 hypotheses from 20 candidates (4 models $\times$ 5 slots), whereas Default reports all 5 from one model, so part of the gain may simply be inference-time scaling (picking the best 5 of 20 rather than keeping 5 of 5), and the two settings are scored by different judge panels.

\paragraph{Paper-level uncertainty and success@5.}
Table~\ref{tab:main}'s $\pm$ values describe dispersion across judges.
To quantify uncertainty from the sampled papers, we recompute Match rates under paper-level bootstrap on the same aligned $n{=}643$ papers, comparing multi-agent to the best fixed top~4 dagger model in each domain (Table~\ref{tab:bootstrap}; full procedure in Appendix~\ref{app:bootstrap}).
Briefly: the resampling unit is a paper (keeping that paper's full set of judged hypotheses); within each domain we draw $n_d$ papers with replacement, recompute domain means, and form lift as the ratio of bootstrap means; 95\% confidence intervals (CIs) are percentile intervals from $B{=}2000$ replications.
On the pooled $n{=}643$ papers, overall lift is $2.4\times$ with 95\% CI $[2.3, 2.6]$ (Table~\ref{tab:bootstrap}, Overall row).
A paper-wise paired comparison against the \emph{per-paper} max among the four dagger top~4 models (a stronger single-model foil; Appendix~\ref{app:signtest}) favors multi-agent on $343$ vs.\ $160$ papers ($140$ ties; two-sided sign test $p<10^{-15}$).
We also report \textbf{success@5}: the fraction of papers for which at least one hypothesis matches among a judge's eligible hypotheses, averaged over eligible judges.
Multi-agent success@5 is $57.1\%$ overall vs.\ $55.1\%$ for the domain-best dagger single; the gap is smaller than for per-hypothesis Match rate, and in Chemistry and Medicine multi-agent success@5 is slightly below the domain-best dagger (Table~\ref{tab:bootstrap}).
The Match lift therefore comes mainly from recovering \emph{more} matching hypotheses on papers that already have at least one hit, not from covering many additional papers.

\begin{table}[t]
\centering
\scriptsize
\caption{Paper-level bootstrap percentile 95\% CIs for Match rate and lift vs.\ the best fixed top~4 dagger model per domain, plus success@5 (Appendix~\ref{app:bootstrap}).
Lift CI is for the ratio of bootstrap means within each domain (Overall uses the pooled $n{=}643$ ratio).
Paired wins use the stronger per-paper max among top~4 dagger models.}
\label{tab:bootstrap}
\begin{tabular}{@{}lrrrrrr@{}}
\toprule
Domain & $n$ & MA Match [CI] & Best$^{\dagger}$ [CI] & Lift [CI] & MA s@5 & Best$^{\dagger}$ s@5 \\
\midrule
ML & 120 & $22.9$ $[17.9,28.4]$ & $9.3$ $[7.3,11.3]$ & $2.5$ $[2.0,3.0]$ & $40.1$ & $38.9$ \\
Astronomy & 85 & $36.5$ $[30.5,44.2]$ & $15.5$ $[12.6,18.6]$ & $2.4$ $[2.0,2.8]$ & $58.9$ & $56.9$ \\
Chemistry & 105 & $38.4$ $[31.9,45.8]$ & $16.8$ $[14.2,19.5]$ & $2.3$ $[2.0,2.6]$ & $59.0$ & $60.6$ \\
Materials & 117 & $40.1$ $[33.9,46.2]$ & $15.4$ $[12.9,18.0]$ & $2.6$ $[2.2,3.0]$ & $63.3$ & $54.4$ \\
Medicine & 78 & $41.6$ $[34.9,49.6]$ & $15.8$ $[13.3,18.5]$ & $2.6$ $[2.3,3.1]$ & $65.0$ & $66.2$ \\
Physics & 138 & $36.4$ $[31.1,41.7]$ & $15.8$ $[13.6,18.2]$ & $2.3$ $[2.0,2.6]$ & $59.4$ & $58.0$ \\
\midrule
Overall & 643 & $35.6$ $[33.0,38.4]$ & $14.6$ $[13.5,15.7]$ & $2.4$ $[2.3,2.6]$ & $57.1$ & $55.1$ \\
\bottomrule
\end{tabular}
\end{table}

\paragraph{Hypothesis length.}
Table~\ref{tab:length} compares English word counts (whitespace tokens) for seed papers (title$+$abstract) and for hypotheses (title$+$summary) under Default top~4 vs.\ multi-agent.
Overall, Default averages $56$ words (median $49$), multi-agent $114$ (median $95$), and seeds $191$ (median $183$).
Relative to Default, multi-agent is roughly $2\times$ longer in every domain; relative to seeds, however, multi-agent is the closer match (Overall means $114$ vs.\ $191$, versus Default's $56$), so the Default shortfall is the more striking gap.
Seed texts are full title$+$abstract units rather than hypothesis cards, so the comparison is descriptive rather than genre-matched; length/presentation remains a possible confounder pending controls that hold hypothesis length fixed.

\begin{table}[t]
\centering
\scriptsize
\caption{Length in English words by domain (whitespace tokens).
\textbf{Seed}: seed-paper title$+$abstract (one value per paper).
\textbf{Default}/\textbf{MA}: hypothesis title$+$summary (Default: all top~4 hypotheses on the aligned papers; MA: five final hypotheses per paper).}
\label{tab:length}
\begin{tabular}{@{}lrrrrrrrr@{}}
\toprule
 & & \multicolumn{2}{c}{Seed} & \multicolumn{2}{c}{Default} & \multicolumn{2}{c}{MA} \\
\cmidrule(lr){3-4}\cmidrule(lr){5-6}\cmidrule(lr){7-8}
Domain & $n$ & mean & median & mean & median & mean & median \\
\midrule
ML & 120 & 182 & 178 & 45 & 40 & 93 & 77 \\
Astronomy & 85 & 204 & 210 & 67 & 58 & 135 & 112 \\
Chemistry & 105 & 166 & 163 & 53 & 48 & 110 & 90 \\
Materials & 117 & 175 & 172 & 56 & 50 & 113 & 95 \\
Medicine & 78 & 257 & 253 & 58 & 53 & 121 & 104 \\
Physics & 138 & 186 & 188 & 58 & 50 & 117 & 101 \\
\midrule
Overall & 643 & 191 & 183 & 56 & 49 & 114 & 95 \\
Avg (domains) & --- & 195 & 194 & 56 & 50 & 115 & 96 \\
\bottomrule
\end{tabular}
\end{table}

\paragraph{Candidate-count bounds from the Default top~4 grid.}
The headline Default vs.\ multi-agent comparison is not candidate-matched: multi-agent selects 5 of 20 hypotheses, whereas Default reports 5 of 5 from one model.
As a zero-compute bound that reuses existing Default top~4 judgments (dagger peer judges), Table~\ref{tab:candbound} evaluates four constructions on the same $4\times5$ grid (Appendix~\ref{app:candbound}).
The pool mean over all 20 hypotheses (B) equals the conditional expectation of selecting one hypothesis uniformly at random in each slot, $\mathbb{E}[A]=B$. Table~\ref{tab:candbound} reports one seeded realization of $A$, so its value need not equal $B$; both lie below the best fixed single model (Overall 12.3\% and 12.6\%, respectively, vs.\ 14.6\%).
An infeasible slot oracle that peeks at Match labels on the unreviewed Default grid (C) reaches 30.8\%, while the reviewed multi-agent champions reach 35.6\% (Overall; $+23.3\%$ over A and $+4.8\%$ over C).
If multi-agent were only selecting among unchanged Default hypotheses under the same dagger scores, one would expect $\mathrm{MA}\le C$; the observed $\mathrm{MA}>C$ in every domain is therefore \emph{not} an oracle violation.
Column C is an upper bound only within the frozen Default $4\times5$ grid scored by dagger peers \emph{with the slot-alignment constraint}, whereas multi-agent may revise candidates during reference-only review and is scored by an origin-recused top~4 panel.
A still stronger Default-grid oracle that ignores slots and keeps the five highest-scoring hypotheses among all 20 (D) reaches 44.5\% Overall, so $\mathrm{C}<\mathrm{MA}<\mathrm{D}$ on the pooled set (and $\mathrm{MA}<D$ in every domain).
Thus multi-agent exceeds slot-wise Default re-picking, yet remains below an unconstrained label peek at the same Default pool---consistent with review$+$Swiss adding value beyond per-slot selection, while leaving headroom relative to an infeasible 5-of-20 Default cherry-pick (and relative to matched same-panel controls still to come).
Judge-panel mismatch remains a caveat, so these bounds still do not replace a matched best-of-20 control with shared review, Swiss, and judge panels.

\begin{table}[t]
\centering
\scriptsize
\caption{Zero-compute bounds on candidate count from the Default top~4 $4\times5$ grid (dagger peer judges).
\textbf{Best$^{\dagger}$}: fixed domain-best top~4 model.
\textbf{A}: one seeded realization of slot-wise random selection (one of four models chosen uniformly in each slot; seed 42).
\textbf{B}: mean Match rate over all 20 hypotheses.
By linearity, $\mathbb{E}[A]=B$ for each paper (Appendix~\ref{app:candbound}), but the reported realization of $A$ need not equal $B$.
\textbf{C}: slot oracle ($\max$ over four models per slot; match-label peeking).
\textbf{D}: unconstrained oracle (mean of the five highest among all 20; match-label peeking, no slot constraint).
\textbf{MA}: multi-agent Match rate (Table~\ref{tab:main}).
Entries are domain means of paper-level rates (\%).
This does \emph{not} replace a matched best-of-20 control.}
\label{tab:candbound}
\begin{tabular}{@{}lrrrrrrr@{}}
\toprule
Domain & $n$ & Best$^{\dagger}$ & A & B & C & D & MA \\
\midrule
ML & 120 & 9.3 & 7.1 & 8.5 & 22.0 & 31.3 & 22.9 \\
Astronomy & 85 & 15.5 & 14.6 & 13.0 & 31.6 & 46.9 & 36.5 \\
Chemistry & 105 & 16.8 & 12.2 & 14.2 & 33.7 & 47.7 & 38.4 \\
Materials & 117 & 15.4 & 13.2 & 13.2 & 32.9 & 45.8 & 40.1 \\
Medicine & 78 & 15.8 & 14.7 & 13.8 & 32.9 & 51.1 & 41.6 \\
Physics & 138 & 15.8 & 13.5 & 13.3 & 32.5 & 47.1 & 36.4 \\
\midrule
Overall & 643 & 14.6 & 12.3 & 12.6 & 30.8 & 44.5 & 35.6 \\
Avg (domains) & --- & 14.8 & 12.5 & 12.7 & 30.9 & 45.0 & 36.0 \\
\bottomrule
\end{tabular}
\end{table}

\subsection{Parametric contamination strata}
\label{sec:contam}
The protocol in Section~\ref{sec:anti-leakage} prevents prompt-time access to the seed; it does not prevent parametric memorization if a seed entered pretraining before a model's knowledge cutoff (KC).
Appendix~\ref{app:contam} therefore stratifies the aligned $n{=}643$ seeds by \emph{earliest public date} (arXiv / Crossref / Semantic Scholar) against each evaluated model's KC.
We treat date $\le$ KC as an \emph{upper bound on corpus reachability}: papers strictly after that bound could not have entered pretraining under the stated KC, and form the cleaner anti-memorization slice.
Month-only KCs are filled to month-end; starred KCs are inferred as release${-}116$ days (the minimum known release$-$KC gap among Claude, GPT, and Gemini).

Most seeds are from 2026 ($498/643{=}77.4\%$), with Chemistry and Medicine almost entirely 2026 (Table~\ref{tab:contam-year-domain}).
The share of papers with date $\le$ KC ranges from $5.1\%$ (Gemini, KC 2025-01-31) to $63.3\%$ (Kimi, inferred KC 2026-03-22; Table~\ref{tab:contam-possible-domain}).

A common split at 2026-03-22 (Kimi's inferred KC) gives $n_{\le}{=}407$ vs.\ $n_{>}{=}236$.
On the later slice, multi-agent Match rates are $19.7\%$ (ML), $33.0\%$ (Astronomy), $31.7\%$ (Chemistry), $31.8\%$ (Materials), $41.6\%$ (Medicine), and $28.7\%$ (Physics), averaging $31.1\%\pm6.5\%$---still well above the best dagger single-model average on the same slice ($12.4\%$).
The vs-best-single lift, averaged across the six domains, remains $2.4\times$ (domain ratios $1.7$--$3.5\times$; Table~\ref{tab:contam-date-split}).
All seven Default models and multi-agent have a positive average $\Delta=\mathrm{Match}(\le)-\mathrm{Match}(>)$ (Default mean $+2.7\%$; multi-agent $+8.5\%$), concentrated in Chemistry, Materials, Physics, and Astronomy; ML is often flat or reversed (Table~\ref{tab:contam-delta-A}).
Splitting instead by each proposer's own KC preserves the sign pattern (Tables~\ref{tab:contam-kc-split} and~\ref{tab:contam-delta-B}).

These $\Delta$ values compare two \emph{disjoint} paper sets of unequal size and different domain/year mix; they are not within-paper changes and should not be read as measuring how much memorization raises Match rate.
Date $\le$ KC also does not imply instance-level storage: generating a similar idea is not proof of memorization, and models trained far beyond capacity are predicted to be near chance on membership of the average training point~\cite{morris2025memorize}.
The tighter observation is therefore the \emph{nonzero} Match on the $>$ slice, which is compatible with bibliography-conditioned inference---and with remaining confounds such as domain mix and hypothesis length---rather than a parametric copy of the seed.

\subsection{Takeaways}
\begin{itemize}
  \item \textbf{Blind idea recovery is hard for single models.} Even the strongest frontier models recover the true idea for only a small fraction of hypotheses under a strict pre-publication cutoff ($3.4$--$15.0\%$ Match rates).
  \item \textbf{Reference-only review + tournament selection helps substantially.} The full pipeline---without extra search---reaches $22.9$--$41.6\%$ absolute Match rates across domains (mean $36.0\%$).
  Relative ratios vs.\ best dagger single fall in a narrow $2.3$--$2.6\times$ band; we report them as descriptive associations pending ablations.
  Paper-level bootstrap supports an overall ratio of $2.4\times$ (95\% CI $[2.3,2.6]$); success@5 rises more modestly ($57.1\%$ vs.\ $55.1\%$).
  \item \textbf{Beyond Default-grid selection.} Multi-agent exceeds the Default $4\times5$ slot oracle C in every domain (Table~\ref{tab:candbound}; Overall $+4.8\%$), which would be impossible if it only re-picked unchanged Default hypotheses under the same dagger scores.
  It remains below the unconstrained Default top-5 oracle D (Overall $35.6\%$ vs.\ $44.5\%$), so $\mathrm{C}<\mathrm{MA}<\mathrm{D}$: the pipeline beats slot-wise Default selection, yet does not saturate an infeasible cherry-pick of any five Default hypotheses.
  Together with the large gap over random/pool baselines A/B, this supports that reference-only review$+$Swiss selection contributes to recovery beyond inference-time expansion of the Default pool---while still leaving a matched same-panel best-of-20 control for future work.
  \item \textbf{The pattern is not confined to possibly-seen papers.} On seeds first public after \mbox{2026-03-22} ($n{=}236$), multi-agent remains in a $19.7$--$41.6\%$ domain band (mean $31.1\%$) with a $2.4\times$ average lift vs.\ the best dagger single on that slice (Section~\ref{sec:contam}).
  Date $\le$ KC bounds corpus reachability, not instance-level memorization; the $\Delta$ between $\le$ and $>$ slices compares disjoint sets of unequal mix, not a within-paper memorization effect.
\end{itemize}

\section{Limitations}

\begin{itemize}
  \item The evaluated LLMs are recent frontier models (or near-frontier) from each provider; seed papers may appear in their pretraining corpora, so recovered ideas could partly reflect memorized content rather than bibliography-conditioned inference.
  Our reference-only protocol and pre-publication citation cutoff reduce \emph{prompt-time} leakage, but cannot rule out parametric contamination (Section~\ref{sec:contam}, Appendix~\ref{app:contam}).
  Earliest-public-date vs.\ knowledge-cutoff is a reachability proxy (some KCs are inferred), not an instance-level membership test.
  \item Match rate depends on LLM judges; we mitigate self-evaluation bias via leave-one-out / origin recusal and report judge variance, but we do not yet report human agreement with the Match rubric (Appendix~\ref{app:judge}).
  \item The headline Default vs.\ multi-agent comparison is not an isolated collaboration effect.
  Multi-agent selects 5 of 20 hypotheses (4 models $\times$ 5 slots), while Default keeps 5 of 5 from one model; part of the observed $2.4\times$ association may therefore reflect inference-time scaling---picking among more candidates---rather than cross-model review or tournament mechanics per se.
  Table~\ref{tab:candbound} gives zero-compute bounds from the Default top~4 grid---including that multi-agent exceeds the Default-grid slot oracle C, which we interpret as supporting review$+$selection beyond pure Default re-picking---but a matched best-of-20 single-model control with shared review/Swiss/judge panels is still planned.
  \item The two settings are also scored by different judge panels (six leave-one-out judges for primary Default rows vs.\ top~4 with origin recusal for multi-agent).
  Table~\ref{tab:main}'s dagger rows partially align the Default top~4 panel, but the comparisons remain not strictly like-for-like.
  \item The top~4 roster is post-selected on the reported Default averages (same papers), so multi-agent evaluates this ensemble rather than a selection policy validated on held-out data.
  \item Multi-agent uses more total compute than a single model; we do not yet normalize for equal token budgets.
  \item Hypothesis ``match'' is a coarse binary; finer-grained graded scoring (e.g., a 1--5 rubric) can be considered.
  \item Multi-agent hypotheses are longer than Default ones but closer in length to seed title$+$abstract than Default is (Overall means $114$ vs.\ $56$ vs.\ seed $191$; Table~\ref{tab:length}); part of the Match lift could still be a length/presentation artifact until length-matched controls are reported.
  \item The reported seed-paper set covers only six domains (ML plus five Nature-family areas); broader scientific coverage remains open.
\end{itemize}

\section{Future Work}

Reconstruction is only the first step.
The experiments show that, under this protocol, the reference-only selection harness raises Match rates well above the single-model Default band, including on papers first public after \mbox{2026-03-22}; relative ratios vs.\ the best dagger single remain an observed association for the full pipeline, pending candidate-matched and equal-compute controls.

The end goal, however, is not reconstruction.
We ultimately aim to \emph{generate} novel and feasible research ideas---the Generation mode of AI-Professor---where models propose forward-looking directions rather than recover known ones.
Reconstruction serves as a controlled stress test of literature understanding and multi-agent coordination under a strict information cutoff; the same harness principles (heterogeneous models, reference-grounded critique, and tournament selection) are intended to transfer to open-ended ideation, where success must be measured by novelty, feasibility, and real-world research impact rather than Match rate against a held-out seed paper.
Immediate next steps include a candidate-matched single-model best-of-20 control, human validation of the Match judge, expanding other domain coverage, and bridging Reconstruction findings into Generation evaluation.

\section{Conclusion}

We presented Reconstruction, a blind bibliography-to-idea recovery benchmark with a strict prompt-time anti-leakage protocol, and showed that a \textbf{multi-agent (top~4)} pipeline---reference-only cross-model review plus Swiss selection---reaches Match rates of $22.9$--$41.6\%$ across six scientific domains, versus a single-model Default regime of $3.4$--$15.0\%$, with an observed $2.4\times$ lift over the best dagger single-model baseline.
The same qualitative pattern holds on papers first public after \mbox{2026-03-22}.

\section*{Acknowledgments}
Implementation is part of the AI-Professor system.

\clearpage
\bibliographystyle{plain}
\bibliography{references}

\clearpage
\appendix
\section{Final Match judge prompt and rubric}
\label{app:judge}

Each hypothesis is judged independently against the held-out seed title/abstract.
The judge model never sees other candidates in the same call and must differ from the proposer (leave-one-out / origin recusal).
For multi-agent scoring, the origin judge's label for a champion is missing and excluded from aggregation; the champion score is the mean of its three eligible non-origin labels.

\paragraph{System prompt (verbatim).}
\begin{quote}
\footnotesize
You judge whether a single candidate research hypothesis matches a held-out seed paper.\\
Rules:\\
- Compare ONLY the supplied title and abstract for the seed paper against the hypothesis title and summary.\\
- Do NOT use external knowledge of the seed paper beyond the supplied title and abstract.\\
- matched=true when the hypothesis describes the same core research idea as the seed paper.\\
- matched=false when unrelated, overly generic, only loosely related, or not about the same research question.\\
- Judge only the one hypothesis provided; do not speculate about other candidates.\\
- rationale: one or two concise English sentences explaining the match decision (what aligns or what diverges).\\
- Respond in English with valid JSON matching the requested schema.
\end{quote}

\paragraph{Binary Match rubric.}
\texttt{matched=true} iff the hypothesis describes the \emph{same core research idea} as the seed (same research question / central claim), not merely the same broad topic.
Borderline guidance used in development:
\begin{itemize}
  \item \textbf{Match:} same problem and essentially the same proposed mechanism/claim, even if wording differs.
  \item \textbf{No match:} same field or overlapping keywords but a different question; a generic restatement of the bibliography without the seed's distinctive claim; or a related but alternate approach that would not be recognized as the seed paper's idea.
\end{itemize}
The user message supplies seed and hypothesis title/abstract blocks and requests JSON
\texttt{\{"matched": true|false, "rationale": "..."\}}.

\paragraph{Borderline examples.}
To make the binary rubric concrete, Table~\ref{tab:borderline} shows one unanimous \textbf{Match} and one unanimous \textbf{No-match} illustration.
The two examples use different domains and different top~4 proposers.
Within each example, the Default and multi-agent (MA) hypotheses share the same seed paper and the same origin proposer.
For each seed we show the title and abstract; for each hypothesis we show the title and summary.
Both candidates come from a high-agreement pool: Default judges unanimous, the three non-origin MA judges unanimous, and the MA hypothesis a Swiss full-sweep slot champion (every match scored $2.0$).

\newpage
{\setlength{\LTcapwidth}{\textwidth}%
\setlength{\tabcolsep}{4pt}
\begin{longtable}{@{}>{\scriptsize}p{1.9cm}>{\scriptsize}p{13.6cm}@{}}
\caption{Borderline Match / No-match examples illustrating the final judge rubric (Appendix~\ref{app:judge}).
Within each example the Default and multi-agent (MA) hypotheses share the same seed paper and origin proposer.
Seeds are shown as title plus abstract; hypotheses as title plus summary.}
\label{tab:borderline}\\
\toprule
\endfirsthead
\multicolumn{2}{@{}l@{}}{\scriptsize\textit{Table~\ref{tab:borderline} (continued).}}\\
\toprule
\endhead
\bottomrule
\endfoot
\bottomrule
\endlastfoot
\multicolumn{2}{@{}l@{}}{\scriptsize\textbf{Match --- ML} (proposer: Kimi-K3)}\\[0.25em]
seed & {\raggedright\textbf{How much do language models memorize?}\par\smallskip We propose a new method for estimating how much a model knows about a datapoint and use it to measure the capacity of modern language models. Prior studies of language model memorization have struggled to disentangle memorization from generalization. We formally separate memorization into two components: unintended memorization, the information a model contains about a specific dataset, and generalization, the information a model contains about the true data-generation process. When we completely eliminate generalization, we can compute the total memorization, which provides an estimate of model capacity: our measurements estimate that GPT-style models have a capacity of approximately 3.6 bits per parameter. We train language models on datasets of increasing size and observe that models memorize until their capacity fills, at which point"grokking"begins, and unintended memorization decreases as models begin to generalize. We train hundreds of transformer language models ranging from 500K to 1.5B parameters and produce a series of scaling laws relating model capacity and data size to membership inference.\par} \\[0.3em]
Default & {\raggedright\textbf{How Many Bits Can a Language Model Memorize? Measuring Raw Memorization Capacity with Structureless Data}\par\smallskip Train transformer language models on uniformly random, structure-free token sequences so that generalization is impossible by construction, then use compression- and information-theoretic accounting to measure the total number of bits the model stores, yielding a clean bits-per-parameter estimate of memorization capacity and a principled separation of memorization from generalization.\par} \\[0.3em]
MA & {\raggedright\textbf{How Many Bits per Parameter? Measuring Stored, Retrievable, and Extractable Memorization of Language Models via Keyed Random-Data Pretraining}\par\smallskip Isolate pure memorization from generalization by pretraining decoder-only transformers on synthetic corpora of key-to-payload records built from uniformly random bitstrings, which are incompressible with high probability, so any predictability the model acquires is memorization by construction. Using a compression-based, description-length definition of memorization, measure three nested quantities---stored bits (code-length reduction given the model), retrievable bits (exact payload recovery given the key), and extractable bits (black-box recovery without keys, following)---as functions of parameter count, data load, epochs, and numerical precision, yielding empirical capacity curves rather than a presumed universal constant.\par} \\[0.3em]
\midrule
\multicolumn{2}{@{}l@{}}{\scriptsize\textbf{No-match --- Physics} (proposer: Claude-Opus-4.8)}\\[0.25em]
seed & {\raggedright\textbf{Emergence of phase coherence in a magnon Bose--Einstein condensate}\par\smallskip The spontaneous emergence of coherence is a defining feature of multibody quantum systems, underlying phenomena from superconductivity to quantum information processing. Although Bose--Einstein condensates provide a unique setting for studying this process, direct observation of how a condensate acquires a coherent global phase has remained challenging. Here we provide evidence of spontaneous phase formation in a magnon Bose--Einstein condensate. Using a phase-referenced detection technique, we track the phase of the condensate coherent state relative to an external reference, revealing how an initially incoherent magnon gas thermalizes and undergoes a spontaneous transition into a coherent quantum state with a well-defined macroscopic phase. This observation provides evidence of spontaneous symmetry breaking in a quasiparticle condensate, confirming a central prediction of Bose--Einstein condensation theory that extends across diverse quantum systems.\par} \\[0.3em]
Default & {\raggedright\textbf{Interference of Two Independently Prepared Room-Temperature Magnon Condensates}\par\smallskip Prepare two spatially separated magnon Bose-Einstein condensates in a YIG film via localized parametric pumping and let them expand and overlap, observing matter-wave-like interference fringes analogous to atomic BEC interference, but at room temperature.\par} \\[0.3em]
MA & {\raggedright\textbf{Thermally Defined Weak Link Between Two Room-Temperature Magnon Condensates: Controlling Josephson Detuning and Barrier Height via Local Laser Heating}\par\smallskip Investigate whether a laser-heated separating region can act as a controllable weak link between two adjacent room-temperature YIG magnon BECs, with local heating tuning the inter-condensate chemical-potential detuning and effective barrier height, and disentangle coherent Josephson population exchange from thermally driven supercurrent outflow and Bogoliubov/second-sound transport.\par} \\[0.3em]
\end{longtable}
}

\section{Swiss tournament mechanics (reconstruction multi-agent)}
\label{app:swiss}

Per hypothesis slot, the four models' slot-$k$ hypotheses form a candidate pool of size $n{=}4$.
All candidates enter Swiss selection (no novelty gate; no external web search---ballots see only the anonymous reference corpus).

\paragraph{Rounds and pairing.}
The number of rounds is $\max(1,n{-}1)$ (here, 3).
Round~0 shuffles candidates and pairs them consecutively; a leftover candidate receives a bye ($+1$ point).
Later rounds rank by current score (then stable IDs) and pair similar scores while avoiding rematches when possible.

\paragraph{Judging rules (conflict-of-interest + order debias).}
For each paired A vs.\ B match under the top~4 roster:
\begin{enumerate}
  \item The proposers of A and of B \emph{recuse}: they cannot judge that match.
  \item The remaining two models \emph{both} serve as judges.
  \item Each judge casts \emph{two} ballots: once with presentation order A/B and once with B/A (swap only the prompt order; the fixed A/B identities of the two candidates do not change), to reduce position bias.
  \item Thus each match yields four ballots in total.
\end{enumerate}
Each ballot independently awards one vote to A or B.
Match points equal vote count$/2$ (the two sides' points always sum to $2.0$): e.g., $4$--$0\mapsto 2.0$--$0$, $3$--$1\mapsto 1.5$--$0.5$, $2$--$2\mapsto 1.0$--$1.0$ (draw).
When a judge's A/B and B/A verdicts disagree, that judge is flagged for position bias in the audit log; ranking still uses the full four-ballot points above.
After all rounds, the highest cumulative Swiss score wins the slot; ties break by aggregate importance then session ID.

\paragraph{Tournament ballot system prompt (verbatim).}
\begin{quote}
\footnotesize
You are a debate judge for Multi-agent Reconstruction in AI Professor.\\
Compare two hypotheses using ONLY the supplied anonymous reference corpus.\\
Pick the hypothesis that better fits the latent research idea implied by that bibliography:\\
- stronger grounding in the given references\\
- clearer capture of the implied problem gap / contribution\\
- more specific and coherent claims supported by references\\
Do NOT prefer novelty for its own sake. Do NOT use external search knowledge.\\
Respond in English with valid JSON:\\
\texttt{\{"winner": "A" or "B", "rationale": "..."\}}
\end{quote}
The user message supplies the anonymous reference corpus and the two candidate hypotheses labeled A/B under the presentation order for that ballot.

\paragraph{Independence from final Match.}
Review and Swiss ballots never see the seed title/abstract.
The final Match judge (Appendix~\ref{app:judge}) is applied only after the five slot champions are assembled, and is independent of prior review/selection calls.

\section{Paper-level bootstrap confidence intervals}
\label{app:bootstrap}

Table~\ref{tab:main}'s $\pm$ values summarize dispersion across judges and do not quantify uncertainty from which papers enter the seed-paper set.
Table~\ref{tab:bootstrap} therefore reports paper-level nonparametric bootstrap percentile intervals~\cite{efron1993bootstrap,davison1997bootstrap}.

\paragraph{Estimands.}
For each aligned paper $i$ in domain $d$, let $m_i^{\mathrm{MA}}$ be the multi-agent Match rate under the weighted eligible-judge aggregation defined in Section~3.1 (equivalently: average each of the $n_s{=}5$ champion slots over its eligible non-origin judges).
Let $m_i^{\mathrm{Best}\dagger}$ be the corresponding rate for the fixed best top~4 dagger proposer in $d$ (other-top~3 judges).
Domain Match rates are means of these paper-level scores; domain lift is the ratio of the two domain means.
The Overall row pools all $n{=}643$ aligned papers and reports the ratio of the two pooled means---our primary lift summary.

\paragraph{Resampling.}
The bootstrap unit is the paper: drawing paper $i$ reuses its complete judged hypothesis set (no hypothesis-level or judge-level resampling), which respects within-paper dependence among the $n_s{=}5$ slots.
For a domain with $n_d$ papers we draw $n_d$ indices uniformly with replacement (\emph{stratified within domain}), recompute both means on the resampled multiset, and set
\[
\mathrm{Lift}^{(b)} \;=\; \frac{\overline{m}^{\mathrm{MA},(b)}}{\overline{m}^{\mathrm{Best}\dagger,(b)}}.
\]
Domain rows in Table~\ref{tab:bootstrap} use this within-domain resampling.
The Overall row instead resamples all 643 papers with replacement and forms the ratio of pooled means.
We use $B{=}2000$ independent replications with a fixed RNG seed for reproducibility.

\paragraph{Percentile 95\% CI.}
For a scalar estimand $\theta$ with bootstrap replicates $\theta^{(1)},\ldots,\theta^{(B)}$, the reported interval is the percentile CI
\[
\bigl[\, Q_{0.025}\bigl(\{\theta^{(b)}\}\bigr),\; Q_{0.975}\bigl(\{\theta^{(b)}\}\bigr) \,\bigr],
\]
i.e., the empirical 2.5\% and 97.5\% quantiles of the bootstrap distribution~\cite{efron1993bootstrap}.
Lift intervals are formed from the $B$ ratios $\mathrm{Lift}^{(b)}$ directly (not by dividing separately obtained Match endpoints).
Point estimates in Table~\ref{tab:bootstrap} are computed on the original (non-resampled) aligned seed-paper set; brackets are bootstrap percentile limits.
Success@5 columns are point estimates only (fraction of papers with at least one Match among a judge's eligible hypotheses, averaged over eligible judges).

\paragraph{Choice of $B$.}
Estimating a standard error typically needs far fewer replications than estimating tail quantiles for a confidence interval~\cite{efron1993bootstrap,davison1997bootstrap}.
Textbook guidance commonly takes on the order of $B{\approx}1000$ replications as adequate for percentile intervals, with larger $B$ reducing Monte Carlo error in the endpoints~\cite{efron1993bootstrap,davison1997bootstrap}; analyses of bootstrap confidence intervals often use on the order of $B{=}2000$ replications for this more delicate task~\cite{diciccio1996bootstrap}.
We therefore set $B{=}2000$.

\newpage
\section{Paper-wise paired comparison and sign test}
\label{app:signtest}

Table~\ref{tab:bootstrap}'s lift compares multi-agent to a single fixed dagger model per domain.
As a stronger, per-paper single-model foil, we also compare multi-agent to the \emph{best-on-that-paper} dagger model, so the baseline is allowed to switch models across papers.

\paragraph{Per-paper foil and win/tie/loss labels.}
For each aligned paper $i$ let $m_i^{\mathrm{MA}}$ be the multi-agent paper-level Match rate (Appendix~\ref{app:bootstrap}) and, for each of the four top~4 dagger proposers $g\in\{1,\ldots,4\}$ (other-top~3 judges), let $m_i^{(g)}$ be that model's paper-level Match rate.
Define the per-paper single-model foil as the maximum over the four models,
\[
m_i^{\max} \;=\; \max_{g\in\{1,\ldots,4\}} m_i^{(g)},
\]
which is a stronger baseline than any fixed model because it may pick a different model on each paper.
We label paper $i$ a multi-agent \emph{win} if $m_i^{\mathrm{MA}} > m_i^{\max}$, a \emph{loss} if $m_i^{\mathrm{MA}} < m_i^{\max}$, and a \emph{tie} if $m_i^{\mathrm{MA}} = m_i^{\max}$.
Over the $n{=}643$ aligned papers this yields $343$ wins, $160$ losses, and $140$ ties ($343{+}160{+}140{=}643$).

\paragraph{Two-sided sign test.}
The sign test asks whether wins and losses are equally likely under the null that multi-agent and the per-paper foil are equally good; ties are uninformative about direction and are discarded, leaving $n_{\mathrm{eff}}=343{+}160=503$ decisive papers.
Let $W$ be the number of wins.
Under $H_0$, $W\sim\mathrm{Binomial}(n_{\mathrm{eff}},\tfrac12)$, and the exact two-sided $p$-value is
\[
p \;=\; \min\!\left\{1,\; 2\sum_{j=0}^{k}\binom{n_{\mathrm{eff}}}{j}2^{-n_{\mathrm{eff}}}\right\},
\qquad k=\min(W,\,n_{\mathrm{eff}}-W).
\]
With $W=343$ and $n_{\mathrm{eff}}=503$ this gives $p\approx2.3\times10^{-16}$, so the paper-level advantage of multi-agent over the per-paper best single model is highly unlikely under chance.
This is a directional consistency test across papers and is distinct from the effect size (lift) and its bootstrap interval in Table~\ref{tab:bootstrap}; it does not by itself quantify how large the per-paper gap is.

\paragraph{Per-domain breakdown.}
Wins/ties/losses (and the same exact two-sided sign-test $p$) by domain are:
ML $50/35/35$ ($p\approx0.13$);
Astronomy $44/22/19$ ($p\approx0.0022$);
Chemistry $56/20/29$ ($p\approx0.0045$);
Materials $67/24/26$ ($p\approx2.5\times10^{-5}$);
Medicine $49/10/19$ ($p\approx3.6\times10^{-4}$);
Physics $77/29/32$ ($p\approx1.9\times10^{-5}$).
The directional advantage is significant at the $0.05$ level in five of six domains; ML is not ($p\approx0.13$).

\newpage
\section{Candidate-count bounds (A/B/C/D) and $\mathbb{E}[A]=B$}
\label{app:candbound}

\paragraph{Setup.}
For each aligned paper, restrict to the four top~4 Default proposers and retain their five slot-aligned hypotheses, yielding scores $r_{k,g}$ for slot $k\in\{1,\ldots,5\}$ and model $g\in\{1,\ldots,4\}$.
Each $r_{k,g}\in[0,1]$ is the fraction of dagger peer judges (the other three top~4 models) that mark the hypothesis as matched.
Let
\[
B \;=\; \frac{1}{20}\sum_{k=1}^{5}\sum_{g=1}^{4} r_{k,g}
\]
be the deterministic mean over the pool of 20 hypotheses (column B in Table~\ref{tab:candbound}).

\paragraph{Slot-wise random selector $A$.}
Draw $G_k$ independently and uniformly from $\{1,2,3,4\}$ and set
\[
A \;=\; \frac{1}{5}\sum_{k=1}^{5} r_{k,G_k}.
\]
Thus $A$ is the paper-level Match rate of keeping one uniformly random Default hypothesis per slot (five hypotheses total).
As a random variable, $A$ is not equal to $B$; only its expectation is.
Table~\ref{tab:candbound} reports one reproducible realization generated with NumPy's PCG64 generator and seed 42.

\paragraph{Proof that $\mathbb{E}[A]=B$.}
Condition on the paper (hence on all $r_{k,g}$). Linearity gives
\begin{align*}
\mathbb{E}[A]
&= \frac{1}{5}\sum_{k=1}^{5}\mathbb{E}[r_{k,G_k}]
= \frac{1}{5}\sum_{k=1}^{5}\sum_{g=1}^{4} r_{k,g}\,\Pr(G_k=g) \\
&= \frac{1}{5}\sum_{k=1}^{5}\sum_{g=1}^{4} r_{k,g}\cdot\frac{1}{4}
= \frac{1}{20}\sum_{k=1}^{5}\sum_{g=1}^{4} r_{k,g}
= B.
\end{align*}
The same identity holds after averaging over papers, so domain/Overall means of $\mathbb{E}[A]$ and of $B$ coincide.
The realized domain/Overall means in column A may differ from column B because they use sampled model choices rather than the conditional expectation.

\paragraph{Slot oracle C and unconstrained oracle D.}
Column C replaces the random draw in each slot by $r_{k,\star}=\max_g r_{k,g}$ and averages over slots (match-label peeking; not deployable).
It is therefore an infeasible upper bound \emph{within} the frozen Default top~4 grid under dagger peer judges \emph{with slot alignment}.
Column D instead sorts all twenty scores $\{r_{k,g}\}$ and averages the five largest, ignoring slots (also match-label peeking; not deployable).
Necessarily $D\ge C$ for each paper, because every slot-wise max is among the twenty candidates.
Multi-agent rates are those reported in Table~\ref{tab:main}.
Empirically $\mathrm{C}<\mathrm{MA}<D$ on the pooled set ($\mathrm{MA}>C$ and $\mathrm{MA}<D$ in all six domains): because multi-agent may revise hypotheses during review and uses a different (origin-recused) judge panel, exceeding C does not contradict the oracle construction; remaining below D leaves headroom relative to an unconstrained Default cherry-pick.

\newpage
\section{Parametric contamination strata (earliest public date vs.\ knowledge cutoff)}
\label{app:contam}

\newlength{\deltanamew}\setlength{\deltanamew}{96pt}
\newlength{\deltacolw}\setlength{\deltacolw}{\dimexpr(\textwidth-84pt-\deltanamew)/7\relax}

The Limitations note that recovered ideas could partly reflect memorized seed content rather than bibliography-conditioned inference.
Section~\ref{sec:contam} states the headline strata; this appendix reports the full observational tables on the aligned $n{=}643$ seeds: earliest public date (arXiv / Crossref / Semantic Scholar) versus each evaluated model's knowledge cutoff (KC).
We treat date $\le$ KC as an \emph{upper bound} on corpus reachability, not on instance-level storage: papers that could not have entered pretraining under that KC bound form the cleaner anti-memorization slice.
Month-only published KCs are filled to month-end; starred KCs are inferred as release${-}116$ days (the minimum known release$-$KC gap among Claude, GPT, and Gemini).

\paragraph{Year $\times$ domain mix.}
Table~\ref{tab:contam-year-domain} shows that most seeds are from 2026 ($498/643{=}77.4\%$), with Chemistry and Medicine almost entirely 2026.

\begin{table}[H]
\centering
\scriptsize
\caption{Year $\times$ domain counts for the aligned $n{=}643$ seeds (earliest public date).
Each cell shows count and share of that domain's row total.}
\label{tab:contam-year-domain}
\begin{tabular}{@{}lccccc@{}}
\toprule
Domain & 2023 & 2024 & 2025 & 2026 & Total \\
\midrule
ML & $0$ ($0.0\%$) & $2$ ($1.7\%$) & $33$ ($27.5\%$) & $85$ ($70.8\%$) & $120$ \\
Astronomy & $0$ ($0.0\%$) & $3$ ($3.5\%$) & $17$ ($20.0\%$) & $65$ ($76.5\%$) & $85$ \\
Chemistry & $0$ ($0.0\%$) & $0$ ($0.0\%$) & $3$ ($2.9\%$) & $102$ ($97.1\%$) & $105$ \\
Materials & $1$ ($0.9\%$) & $6$ ($5.1\%$) & $11$ ($9.4\%$) & $99$ ($84.6\%$) & $117$ \\
Medicine & $0$ ($0.0\%$) & $0$ ($0.0\%$) & $1$ ($1.3\%$) & $77$ ($98.7\%$) & $78$ \\
Physics & $1$ ($0.7\%$) & $16$ ($11.6\%$) & $51$ ($37.0\%$) & $70$ ($50.7\%$) & $138$ \\
\midrule
Total & $2$ ($0.3\%$) & $27$ ($4.2\%$) & $116$ ($18.0\%$) & $498$ ($77.4\%$) & $643$ \\
\bottomrule
\end{tabular}
\end{table}

\paragraph{Possible papers by domain (date $\le$ KC).}
Table~\ref{tab:contam-possible-domain} gives per-model candidate counts by domain.
The possible share of all $643$ seeds ranges from $5.1\%$ (Gemini KC 2025-01-31) to $63.3\%$ (Kimi KC 2026-03-22$^{*}$).

\begin{table}[H]
\centering
\scriptsize
\caption{Possible parametric-contamination candidates by domain: count of aligned seeds whose earliest public date is on or before the model knowledge cutoff (date $\le$ KC).
Month-only KCs are filled to month-end.
Starred KCs are inferred as release${-}116$ days (the minimum known release$-$KC gap among Claude / GPT / Gemini).
The last column is the share of all $643$ seeds.}
\label{tab:contam-possible-domain}
\begin{tabular}{@{}llccccccc@{}}
\toprule
Model & KC & ML & Astro & Chem & Mat & Med & Phys & Possible \% \\
\midrule
Gemini 3.1-Pro-Preview & 2025-01-31 & $2$ & $4$ & $0$ & $7$ & $0$ & $20$ & $5.1\%$ \\
DeepSeek-V4-Pro & 2025-12-29$^{*}$ & $35$ & $20$ & $3$ & $18$ & $1$ & $68$ & $22.6\%$ \\
Qwen3.7-Max & 2026-01-25$^{*}$ & $40$ & $33$ & $21$ & $32$ & $29$ & $76$ & $35.9\%$ \\
Claude-Opus-4.8 & 2026-01-31 & $44$ & $34$ & $24$ & $44$ & $33$ & $81$ & $40.4\%$ \\
GPT-5.6-Sol-Pro & 2026-02-16 & $67$ & $39$ & $28$ & $47$ & $42$ & $87$ & $48.2\%$ \\
GLM-5.2 & 2026-02-20$^{*}$ & $69$ & $39$ & $30$ & $51$ & $44$ & $88$ & $49.9\%$ \\
Kimi-K3 & 2026-03-22$^{*}$ & $85$ & $51$ & $45$ & $72$ & $58$ & $96$ & $63.3\%$ \\
\bottomrule
\end{tabular}
\end{table}

\paragraph{Fixed cut at 2026-03-22.}
As a common paper split for all models (coinciding with Kimi's inferred KC), Table~\ref{tab:contam-date-split} recomputes Table~\ref{tab:main}-style Match rates on $\le$2026-03-22 ($n{=}407$) vs.\ $>$2026-03-22 ($n{=}236$).
Table~\ref{tab:contam-delta-A} reports the per-domain gap $\Delta=\mathrm{Match}(\le)-\mathrm{Match}(>)$.
All seven Default models and multi-agent have positive average $\Delta$ (Default mean $+2.7\%$; multi-agent $+8.5\%$); dagger rows agree in sign with their Default counterparts.
The positive gap is concentrated in Chemistry, Materials, Physics, and Astronomy; ML is often flat or reversed.

{\setlength{\LTcapwidth}{\textwidth}%
\setlength{\tabcolsep}{3.5pt}
\begin{longtable}{@{}>{\scriptsize}l*{7}{>{\scriptsize}c}@{}}
\caption{Match rates (\%) stratified at earliest public date 2026-03-22.
Per model: all ($n{=}643$), $\le$ ($n{=}407$), $>$ ($n{=}236$).
Aggregation matches Table~\ref{tab:main}.
\texttt{vs best single} is multi-agent divided by the best dagger score in that column within the same date stratum.}
\label{tab:contam-date-split}\\
\toprule
Model / split & ML & Astro & Chem & Mat & Med & Phys & Avg \\
\midrule
\endfirsthead
\multicolumn{8}{@{}l@{}}{\scriptsize\textit{Table~\ref{tab:contam-date-split} (continued).}}\\
\toprule
Model / split & ML & Astro & Chem & Mat & Med & Phys & Avg \\
\midrule
\endhead
\bottomrule
\endfoot
\bottomrule
\endlastfoot
Claude-Opus-4.8 (all)
  & $8.2{\pm}2.4$ & $14.7{\pm}3.1$ & $14.0{\pm}5.3$ & $14.0{\pm}3.7$ & $14.2{\pm}3.0$ & $15.0{\pm}3.2$ & $13.3{\pm}2.3$ \\
Claude-Opus-4.8 ($\le$2026-03-22)
  & $8.0{\pm}2.4$ & $15.5{\pm}3.0$ & $16.7{\pm}5.8$ & $15.8{\pm}3.4$ & $15.5{\pm}3.4$ & $16.2{\pm}3.2$ & $14.6{\pm}3.0$ \\
Claude-Opus-4.8 ($>$2026-03-22)
  & $8.6{\pm}2.7$ & $13.4{\pm}3.5$ & $12.0{\pm}5.0$ & $11.0{\pm}4.3$ & $10.5{\pm}2.1$ & $12.3{\pm}3.6$ & $11.3{\pm}1.5$ \\
GPT-5.6-Sol-Pro (all)
  & $7.6{\pm}1.3$ & $12.9{\pm}2.6$ & $15.0{\pm}3.8$ & $13.8{\pm}2.9$ & $14.6{\pm}2.7$ & $12.7{\pm}2.1$ & $12.8{\pm}2.4$ \\
GPT-5.6-Sol-Pro ($\le$2026-03-22)
  & $7.7{\pm}1.6$ & $14.4{\pm}3.2$ & $18.4{\pm}4.8$ & $14.8{\pm}3.6$ & $16.0{\pm}2.8$ & $13.8{\pm}2.1$ & $14.2{\pm}3.2$ \\
GPT-5.6-Sol-Pro ($>$2026-03-22)
  & $7.3{\pm}0.8$ & $10.7{\pm}1.8$ & $12.4{\pm}3.1$ & $12.2{\pm}2.2$ & $10.7{\pm}2.5$ & $10.0{\pm}2.1$ & $10.6{\pm}1.7$ \\
Kimi-K3 (all)
  & $7.9{\pm}1.9$ & $10.7{\pm}2.6$ & $9.8{\pm}2.6$ & $10.1{\pm}2.2$ & $11.1{\pm}1.9$ & $10.7{\pm}2.4$ & $10.0{\pm}1.0$ \\
Kimi-K3 ($\le$2026-03-22)
  & $8.8{\pm}2.1$ & $10.9{\pm}2.9$ & $11.9{\pm}3.7$ & $11.6{\pm}2.4$ & $11.1{\pm}2.1$ & $11.2{\pm}2.3$ & $10.9{\pm}1.0$ \\
Kimi-K3 ($>$2026-03-22)
  & $6.0{\pm}1.5$ & $10.3{\pm}2.4$ & $8.3{\pm}1.9$ & $7.6{\pm}2.4$ & $11.0{\pm}1.5$ & $9.4{\pm}2.8$ & $8.8{\pm}1.7$ \\
GLM-5.2 (all)
  & $7.2{\pm}1.2$ & $9.5{\pm}2.8$ & $9.8{\pm}3.0$ & $9.2{\pm}2.3$ & $10.1{\pm}2.0$ & $10.2{\pm}2.5$ & $9.3{\pm}1.0$ \\
GLM-5.2 ($\le$2026-03-22)
  & $6.3{\pm}0.9$ & $9.5{\pm}3.0$ & $13.6{\pm}4.6$ & $10.7{\pm}2.5$ & $11.3{\pm}1.8$ & $11.3{\pm}2.9$ & $10.5{\pm}2.2$ \\
GLM-5.2 ($>$2026-03-22)
  & $9.3{\pm}2.3$ & $9.5{\pm}2.8$ & $6.9{\pm}1.9$ & $6.9{\pm}2.1$ & $6.8{\pm}2.7$ & $7.5{\pm}1.8$ & $7.8{\pm}1.1$ \\
Gemini 3.1-Pro-Preview (all)
  & $5.9{\pm}0.6$ & $8.9{\pm}1.6$ & $10.2{\pm}2.6$ & $8.4{\pm}2.2$ & $10.5{\pm}1.8$ & $9.2{\pm}1.7$ & $8.9{\pm}1.5$ \\
Gemini 3.1-Pro-Preview ($\le$2026-03-22)
  & $5.7{\pm}0.6$ & $9.3{\pm}2.1$ & $13.6{\pm}3.3$ & $9.7{\pm}2.8$ & $10.9{\pm}2.2$ & $9.4{\pm}1.6$ & $9.8{\pm}2.4$ \\
Gemini 3.1-Pro-Preview ($>$2026-03-22)
  & $6.6{\pm}0.7$ & $8.2{\pm}1.4$ & $7.7{\pm}2.4$ & $6.3{\pm}1.4$ & $9.3{\pm}0.9$ & $8.7{\pm}2.2$ & $7.8{\pm}1.1$ \\
DeepSeek-V4-Pro (all)
  & $3.4{\pm}1.3$ & $6.7{\pm}2.5$ & $6.6{\pm}2.8$ & $7.2{\pm}3.0$ & $6.9{\pm}2.2$ & $7.0{\pm}2.6$ & $6.3{\pm}1.3$ \\
DeepSeek-V4-Pro ($\le$2026-03-22)
  & $3.2{\pm}1.1$ & $8.1{\pm}3.0$ & $7.6{\pm}3.9$ & $8.9{\pm}3.4$ & $7.4{\pm}2.5$ & $7.9{\pm}2.7$ & $7.2{\pm}1.8$ \\
DeepSeek-V4-Pro ($>$2026-03-22)
  & $4.0{\pm}1.6$ & $4.5{\pm}1.8$ & $5.8{\pm}2.1$ & $4.3{\pm}2.4$ & $5.3{\pm}1.6$ & $5.1{\pm}2.3$ & $4.8{\pm}0.6$ \\
Qwen3.7-Max (all)
  & $4.6{\pm}1.3$ & $5.3{\pm}2.2$ & $8.4{\pm}2.5$ & $5.9{\pm}2.4$ & $4.8{\pm}1.8$ & $6.7{\pm}1.8$ & $5.9{\pm}1.3$ \\
Qwen3.7-Max ($\le$2026-03-22)
  & $3.8{\pm}1.4$ & $6.1{\pm}2.5$ & $11.3{\pm}2.8$ & $7.2{\pm}2.8$ & $5.9{\pm}1.8$ & $8.0{\pm}1.7$ & $7.1{\pm}2.3$ \\
Qwen3.7-Max ($>$2026-03-22)
  & $6.3{\pm}1.1$ & $3.9{\pm}2.0$ & $6.3{\pm}2.3$ & $3.8{\pm}1.9$ & $1.7{\pm}1.9$ & $3.6{\pm}2.2$ & $4.3{\pm}1.6$ \\
\midrule
\multicolumn{8}{@{}l@{}}{\scriptsize\textit{Top~4 proposers, judged only by the other top~3 models}} \\
Claude-Opus-4.8$^{\dagger}$ (all)
  & $8.8{\pm}3.0$ & $15.5{\pm}4.0$ & $16.3{\pm}6.0$ & $15.4{\pm}4.6$ & $15.3{\pm}3.3$ & $15.8{\pm}3.8$ & $14.5{\pm}2.6$ \\
Claude-Opus-4.8$^{\dagger}$ ($\le$2026-03-22)
  & $8.6{\pm}3.0$ & $16.7{\pm}3.6$ & $19.3{\pm}5.9$ & $17.4{\pm}4.0$ & $16.7{\pm}3.8$ & $16.8{\pm}4.0$ & $15.9{\pm}3.4$ \\
Claude-Opus-4.8$^{\dagger}$ ($>$2026-03-22)
  & $9.3{\pm}3.0$ & $13.7{\pm}4.6$ & $14.1{\pm}6.0$ & $12.2{\pm}5.6$ & $11.3{\pm}2.6$ & $13.7{\pm}3.9$ & $12.4{\pm}1.7$ \\
GPT-5.6-Sol-Pro$^{\dagger}$ (all)
  & $7.9{\pm}1.6$ & $13.6{\pm}3.3$ & $16.8{\pm}4.3$ & $15.2{\pm}2.8$ & $15.8{\pm}2.7$ & $13.9{\pm}2.2$ & $13.9{\pm}2.9$ \\
GPT-5.6-Sol-Pro$^{\dagger}$ ($\le$2026-03-22)
  & $8.0{\pm}1.9$ & $15.2{\pm}4.3$ & $20.7{\pm}5.1$ & $16.2{\pm}3.6$ & $17.1{\pm}2.9$ & $15.1{\pm}2.1$ & $15.4{\pm}3.8$ \\
GPT-5.6-Sol-Pro$^{\dagger}$ ($>$2026-03-22)
  & $7.6{\pm}0.7$ & $11.4{\pm}1.9$ & $13.8{\pm}3.8$ & $13.5{\pm}1.7$ & $12.0{\pm}2.2$ & $10.9{\pm}2.4$ & $11.5{\pm}2.0$ \\
Kimi-K3$^{\dagger}$ (all)
  & $9.3{\pm}1.6$ & $11.1{\pm}3.0$ & $11.4{\pm}2.9$ & $11.3{\pm}1.6$ & $12.1{\pm}2.2$ & $11.4{\pm}2.8$ & $11.1{\pm}0.9$ \\
Kimi-K3$^{\dagger}$ ($\le$2026-03-22)
  & $10.1{\pm}1.8$ & $11.0{\pm}3.6$ & $14.1{\pm}4.1$ & $13.4{\pm}1.8$ & $12.1{\pm}2.5$ & $11.8{\pm}2.6$ & $12.1{\pm}1.4$ \\
Kimi-K3$^{\dagger}$ ($>$2026-03-22)
  & $7.2{\pm}1.2$ & $11.4{\pm}2.4$ & $9.4{\pm}2.1$ & $7.8{\pm}1.5$ & $12.0{\pm}1.6$ & $10.5{\pm}3.2$ & $9.7{\pm}1.7$ \\
GLM-5.2$^{\dagger}$ (all)
  & $8.1{\pm}0.3$ & $11.7{\pm}2.1$ & $12.3{\pm}2.2$ & $11.2{\pm}1.4$ & $11.9{\pm}0.9$ & $12.2{\pm}1.9$ & $11.2{\pm}1.5$ \\
GLM-5.2$^{\dagger}$ ($\le$2026-03-22)
  & $6.7{\pm}0.2$ & $12.0{\pm}2.0$ & $17.3{\pm}3.2$ & $12.8{\pm}1.6$ & $12.8{\pm}0.8$ & $13.8{\pm}2.0$ & $12.6{\pm}3.1$ \\
GLM-5.2$^{\dagger}$ ($>$2026-03-22)
  & $11.4{\pm}0.8$ & $11.2{\pm}2.5$ & $8.6{\pm}1.5$ & $8.6{\pm}1.5$ & $9.3{\pm}1.2$ & $8.6{\pm}1.7$ & $9.6{\pm}1.2$ \\
\midrule
multi-agent (top 4) (all)
  & $22.9{\pm}6.0$ & $36.5{\pm}11.4$ & $38.4{\pm}8.5$ & $40.1{\pm}10.3$ & $41.6{\pm}10.3$ & $36.4{\pm}10.4$ & $36.0{\pm}6.1$ \\
multi-agent (top 4) ($\le$2026-03-22)
  & $24.3{\pm}6.4$ & $38.9{\pm}12.1$ & $47.4{\pm}9.7$ & $45.4{\pm}11.0$ & $41.6{\pm}10.1$ & $39.8{\pm}10.2$ & $39.6{\pm}7.5$ \\
multi-agent (top 4) ($>$2026-03-22)
  & $19.7{\pm}5.0$ & $33.0{\pm}10.6$ & $31.7{\pm}7.6$ & $31.8{\pm}9.7$ & $41.6{\pm}11.3$ & $28.7{\pm}10.9$ & $31.1{\pm}6.5$ \\
\midrule
vs best single (all)
  & $2.5\times$ & $2.4\times$ & $2.3\times$ & $2.6\times$ & $2.6\times$ & $2.3\times$ & $2.4\times$ \\
vs best single ($\le$)
  & $2.4\times$ & $2.3\times$ & $2.3\times$ & $2.6\times$ & $2.4\times$ & $2.4\times$ & $2.4\times$ \\
vs best single ($>$)
  & $1.7\times$ & $2.4\times$ & $2.2\times$ & $2.4\times$ & $3.5\times$ & $2.1\times$ & $2.4\times$ \\
\end{longtable}
}

{\setlength{\LTcapwidth}{\textwidth}%
\begin{longtable}{@{}>{\scriptsize}l*{7}{>{\scriptsize}c}@{}}
\caption{Per-domain $\Delta = \mathrm{Match}(\le)-\mathrm{Match}(>)$ (\%) for the fixed cut 2026-03-22 (Table~\ref{tab:contam-date-split}).
Positive $\Delta$ means the $\le$ side scores higher.
\emph{Caution:} the two sides are \emph{disjoint} paper sets of unequal size ($n_{\le}{=}407$ vs.\ $n_{>}{=}236$) with different domain and year composition, so $\Delta$ is a difference between two independent subset means, not a within-paper change; per-domain cells rest on even smaller subsets.}
\label{tab:contam-delta-A}\\
\toprule
\makebox[\deltanamew][l]{Model} & \makebox[\deltacolw]{ML} & \makebox[\deltacolw]{Astro} & \makebox[\deltacolw]{Chem} & \makebox[\deltacolw]{Mat} & \makebox[\deltacolw]{Med} & \makebox[\deltacolw]{Phys} & \makebox[\deltacolw]{Avg} \\
\midrule
\endfirsthead
\multicolumn{8}{@{}l@{}}{\scriptsize\textit{Table~\ref{tab:contam-delta-A} (continued).}}\\
\toprule
\makebox[\deltanamew][l]{Model} & \makebox[\deltacolw]{ML} & \makebox[\deltacolw]{Astro} & \makebox[\deltacolw]{Chem} & \makebox[\deltacolw]{Mat} & \makebox[\deltacolw]{Med} & \makebox[\deltacolw]{Phys} & \makebox[\deltacolw]{Avg} \\
\midrule
\endhead
\bottomrule
\endfoot
\bottomrule
\endlastfoot
Claude-Opus-4.8 & $-0.6$ & $+2.1$ & $+4.7$ & $+4.9$ & $+5.0$ & $+3.9$ & $+3.3$ \\
GPT-5.6-Sol-Pro & $+0.4$ & $+3.7$ & $+5.9$ & $+2.7$ & $+5.3$ & $+3.8$ & $+3.6$ \\
Kimi-K3 & $+2.8$ & $+0.6$ & $+3.7$ & $+4.1$ & $+0.1$ & $+1.8$ & $+2.2$ \\
GLM-5.2 & $-3.0$ & $0.0$ & $+6.7$ & $+3.8$ & $+4.4$ & $+3.8$ & $+2.6$ \\
Gemini 3.1-Pro-Preview & $-0.9$ & $+1.1$ & $+6.0$ & $+3.4$ & $+1.5$ & $+0.7$ & $+2.0$ \\
DeepSeek-V4-Pro & $-0.8$ & $+3.6$ & $+1.8$ & $+4.6$ & $+2.1$ & $+2.8$ & $+2.4$ \\
Qwen3.7-Max & $-2.5$ & $+2.2$ & $+5.0$ & $+3.4$ & $+4.2$ & $+4.4$ & $+2.8$ \\
\midrule
\multicolumn{8}{@{}l@{}}{\scriptsize\textit{Top~4 proposers, judged only by the other top~3 models}} \\
Claude-Opus-4.8$^{\dagger}$ & $-0.7$ & $+3.0$ & $+5.2$ & $+5.3$ & $+5.3$ & $+3.2$ & $+3.5$ \\
GPT-5.6-Sol-Pro$^{\dagger}$ & $+0.4$ & $+3.8$ & $+7.0$ & $+2.7$ & $+5.1$ & $+4.2$ & $+3.9$ \\
Kimi-K3$^{\dagger}$ & $+2.9$ & $-0.4$ & $+4.6$ & $+5.6$ & $+0.1$ & $+1.3$ & $+2.3$ \\
GLM-5.2$^{\dagger}$ & $-4.8$ & $+0.8$ & $+8.8$ & $+4.2$ & $+3.4$ & $+5.2$ & $+2.9$ \\
\midrule
\rowcolor{magenta!8}
multi-agent (top 4) & $+4.6$ & $+6.0$ & $+15.7$ & $+13.6$ & $0.0$ & $+11.1$ & $+8.5$ \\
\end{longtable}
}

\paragraph{Split by each model's KC.}
Table~\ref{tab:contam-kc-split} repeats the Match-rate stratification under each proposer's own KC.
Default and dagger $\le$KC/$>$KC rows therefore do \emph{not} share a common paper split: Claude's cut is 2026-01-31, Kimi's is 2026-03-22, and so on.
Multi-agent assigns each paper once using the \emph{latest} KC among its five Swiss winners' origin proposers, so a paper is on the $>$ side only if its earliest public date is after every winner's KC ($n_{\le}{=}406$, $n_{>}{=}237$).
\texttt{vs best single} on those sides is correspondingly not the common-split lift of Table~\ref{tab:contam-date-split}: it compares this max-origin multi-agent split to the best dagger cell in the same column, and those dagger cells rest on heterogeneous subsets.
Table~\ref{tab:contam-delta-B} reports the corresponding per-domain $\Delta$.
Signs remain consistent with the fixed cut; Gemini and DeepSeek average $\Delta$ shrink under their earlier KCs.
Empty cells (---) mark domains with $n{=}0$ on a side (Gemini Chemistry/Medicine on $\le$KC).

{\setlength{\LTcapwidth}{\textwidth}%
\setlength{\tabcolsep}{2.6pt}
\begin{longtable}{@{}>{\scriptsize}l*{7}{>{\scriptsize}c}@{}}
\caption{Match rates (\%) stratified by each proposer's knowledge cutoff (KC).
Default/dagger $\le$KC / $>$KC rows use that proposer's KC at the paper level.
Multi-agent assigns each paper uniquely using the \emph{latest} KC among its five Swiss winners' origin proposers (conservative $>$ set): $n_{\le}{=}406$, $n_{>}{=}237$.
\texttt{vs best single} uses the best dagger score in the same stratum; the $\le$KC/$>$KC lift rows are not a shared paper split.}
\label{tab:contam-kc-split}\\
\toprule
Model / split & ML & Astro & Chem & Mat & Med & Phys & Avg \\
\midrule
\endfirsthead
\multicolumn{8}{@{}l@{}}{\scriptsize\textit{Table~\ref{tab:contam-kc-split} (continued).}}\\
\toprule
Model / split & ML & Astro & Chem & Mat & Med & Phys & Avg \\
\midrule
\endhead
\bottomrule
\endfoot
\bottomrule
\endlastfoot
Claude-Opus-4.8 (all)
  & $8.2{\pm}2.4$ & $14.7{\pm}3.1$ & $14.0{\pm}5.3$ & $14.0{\pm}3.7$ & $14.2{\pm}3.0$ & $15.0{\pm}3.2$ & $13.3{\pm}2.3$ \\
Claude-Opus-4.8 ($\le$2026-01-31) [$n$=260]
  & $8.3{\pm}2.6$ & $15.4{\pm}3.0$ & $18.2{\pm}4.3$ & $18.5{\pm}3.1$ & $15.2{\pm}3.8$ & $16.9{\pm}3.4$ & $15.4{\pm}3.4$ \\
Claude-Opus-4.8 ($>$2026-01-31) [$n$=383]
  & $8.1{\pm}2.3$ & $14.2{\pm}3.5$ & $12.8{\pm}5.6$ & $11.2{\pm}4.2$ & $13.5{\pm}2.4$ & $12.3{\pm}3.2$ & $12.0{\pm}2.0$ \\
GPT-5.6-Sol-Pro (all)
  & $7.6{\pm}1.3$ & $12.9{\pm}2.6$ & $15.0{\pm}3.8$ & $13.8{\pm}2.9$ & $14.6{\pm}2.7$ & $12.7{\pm}2.1$ & $12.8{\pm}2.4$ \\
GPT-5.6-Sol-Pro ($\le$2026-02-16) [$n$=310]
  & $7.9{\pm}1.7$ & $15.6{\pm}2.9$ & $21.4{\pm}4.5$ & $14.8{\pm}3.6$ & $14.7{\pm}2.8$ & $14.6{\pm}2.2$ & $14.8{\pm}3.9$ \\
GPT-5.6-Sol-Pro ($>$2026-02-16) [$n$=333]
  & $7.2{\pm}0.9$ & $10.6{\pm}2.5$ & $12.6{\pm}3.6$ & $13.1{\pm}2.6$ & $14.5{\pm}2.7$ & $9.4{\pm}2.1$ & $11.2{\pm}2.5$ \\
Kimi-K3 (all)
  & $7.9{\pm}1.9$ & $10.7{\pm}2.6$ & $9.8{\pm}2.6$ & $10.1{\pm}2.2$ & $11.1{\pm}1.9$ & $10.7{\pm}2.4$ & $10.0{\pm}1.0$ \\
Kimi-K3 ($\le$2026-03-22) [$n$=407]
  & $8.8{\pm}2.1$ & $10.9{\pm}2.9$ & $11.9{\pm}3.7$ & $11.6{\pm}2.4$ & $11.1{\pm}2.1$ & $11.2{\pm}2.3$ & $10.9{\pm}1.0$ \\
Kimi-K3 ($>$2026-03-22) [$n$=236]
  & $6.0{\pm}1.5$ & $10.3{\pm}2.4$ & $8.3{\pm}1.9$ & $7.6{\pm}2.4$ & $11.0{\pm}1.5$ & $9.4{\pm}2.8$ & $8.8{\pm}1.7$ \\
GLM-5.2 (all)
  & $7.2{\pm}1.2$ & $9.5{\pm}2.8$ & $9.8{\pm}3.0$ & $9.2{\pm}2.3$ & $10.1{\pm}2.0$ & $10.2{\pm}2.5$ & $9.3{\pm}1.0$ \\
GLM-5.2 ($\le$2026-02-20) [$n$=321]
  & $6.6{\pm}0.9$ & $10.8{\pm}3.2$ & $14.4{\pm}4.0$ & $10.8{\pm}2.3$ & $11.5{\pm}1.8$ & $11.9{\pm}3.2$ & $11.0{\pm}2.3$ \\
GLM-5.2 ($>$2026-02-20) [$n$=322]
  & $8.0{\pm}1.9$ & $8.4{\pm}2.6$ & $8.0{\pm}2.7$ & $7.9{\pm}2.3$ & $8.3{\pm}2.3$ & $7.2{\pm}1.5$ & $8.0{\pm}0.4$ \\
Gemini 3.1-Pro-Preview (all)
  & $5.9{\pm}0.6$ & $8.9{\pm}1.6$ & $10.2{\pm}2.6$ & $8.4{\pm}2.2$ & $10.5{\pm}1.8$ & $9.2{\pm}1.7$ & $8.9{\pm}1.5$ \\
Gemini 3.1-Pro-Preview ($\le$2025-01-31) [$n$=33]
  & $0.0{\pm}0.0$ & $17.5{\pm}3.8$ & --- & $9.5{\pm}4.3$ & --- & $10.3{\pm}2.9$ & $9.3{\pm}6.2$ \\
Gemini 3.1-Pro-Preview ($>$2025-01-31) [$n$=610]
  & $6.0{\pm}0.6$ & $8.5{\pm}1.6$ & $10.2{\pm}2.6$ & $8.3{\pm}2.1$ & $10.5{\pm}1.8$ & $9.0{\pm}1.6$ & $8.8{\pm}1.5$ \\
DeepSeek-V4-Pro (all)
  & $3.4{\pm}1.3$ & $6.7{\pm}2.5$ & $6.6{\pm}2.8$ & $7.2{\pm}3.0$ & $6.9{\pm}2.2$ & $7.0{\pm}2.6$ & $6.3{\pm}1.3$ \\
DeepSeek-V4-Pro ($\le$2025-12-29) [$n$=145]
  & $3.2{\pm}0.6$ & $10.2{\pm}2.7$ & $11.1{\pm}3.1$ & $8.9{\pm}3.8$ & $0.0{\pm}0.0$ & $7.7{\pm}2.6$ & $6.8{\pm}4.0$ \\
DeepSeek-V4-Pro ($>$2025-12-29) [$n$=498]
  & $3.5{\pm}1.6$ & $5.6{\pm}2.6$ & $6.4{\pm}2.8$ & $6.8{\pm}3.0$ & $7.0{\pm}2.3$ & $6.4{\pm}2.5$ & $6.0{\pm}1.2$ \\
Qwen3.7-Max (all)
  & $4.6{\pm}1.3$ & $5.3{\pm}2.2$ & $8.4{\pm}2.5$ & $5.9{\pm}2.4$ & $4.8{\pm}1.8$ & $6.7{\pm}1.8$ & $5.9{\pm}1.3$ \\
Qwen3.7-Max ($\le$2026-01-25) [$n$=231]
  & $2.6{\pm}1.3$ & $6.5{\pm}3.2$ & $11.6{\pm}1.3$ & $8.4{\pm}2.4$ & $5.2{\pm}1.5$ & $8.3{\pm}1.8$ & $7.1{\pm}2.8$ \\
Qwen3.7-Max ($>$2026-01-25) [$n$=412]
  & $5.5{\pm}1.6$ & $4.5{\pm}1.8$ & $7.7{\pm}2.9$ & $4.9{\pm}2.5$ & $4.6{\pm}2.0$ & $4.7{\pm}1.9$ & $5.3{\pm}1.1$ \\
\midrule
\multicolumn{8}{@{}l@{}}{\scriptsize\textit{Top~4 proposers, judged only by the other top~3 models}} \\
Claude-Opus-4.8$^{\dagger}$ (all)
  & $8.8{\pm}3.0$ & $15.5{\pm}4.0$ & $16.3{\pm}6.0$ & $15.4{\pm}4.6$ & $15.3{\pm}3.3$ & $15.8{\pm}3.8$ & $14.5{\pm}2.6$ \\
Claude-Opus-4.8$^{\dagger}$ ($\le$2026-01-31) [$n$=260]
  & $9.2{\pm}3.2$ & $17.1{\pm}3.4$ & $20.0{\pm}4.5$ & $19.6{\pm}3.5$ & $17.0{\pm}3.9$ & $17.8{\pm}4.2$ & $16.8{\pm}3.6$ \\
Claude-Opus-4.8$^{\dagger}$ ($>$2026-01-31) [$n$=383]
  & $8.6{\pm}2.9$ & $14.5{\pm}4.5$ & $15.2{\pm}6.5$ & $12.9{\pm}5.3$ & $14.1{\pm}2.9$ & $13.1{\pm}3.7$ & $13.1{\pm}2.1$ \\
GPT-5.6-Sol-Pro$^{\dagger}$ (all)
  & $7.9{\pm}1.6$ & $13.6{\pm}3.3$ & $16.8{\pm}4.3$ & $15.2{\pm}2.8$ & $15.8{\pm}2.7$ & $13.9{\pm}2.2$ & $13.9{\pm}2.9$ \\
GPT-5.6-Sol-Pro$^{\dagger}$ ($\le$2026-02-16) [$n$=310]
  & $8.1{\pm}2.1$ & $16.2{\pm}3.9$ & $22.9{\pm}5.1$ & $16.2{\pm}4.0$ & $16.0{\pm}2.9$ & $15.9{\pm}2.2$ & $15.9{\pm}4.3$ \\
GPT-5.6-Sol-Pro$^{\dagger}$ ($>$2026-02-16) [$n$=333]
  & $7.7{\pm}0.9$ & $11.4{\pm}3.0$ & $14.6{\pm}4.1$ & $14.5{\pm}2.2$ & $15.6{\pm}2.7$ & $10.5{\pm}2.4$ & $12.4{\pm}2.8$ \\
Kimi-K3$^{\dagger}$ (all)
  & $9.3{\pm}1.6$ & $11.1{\pm}3.0$ & $11.4{\pm}2.9$ & $11.3{\pm}1.6$ & $12.1{\pm}2.2$ & $11.4{\pm}2.8$ & $11.1{\pm}0.9$ \\
Kimi-K3$^{\dagger}$ ($\le$2026-03-22) [$n$=407]
  & $10.1{\pm}1.8$ & $11.0{\pm}3.6$ & $14.1{\pm}4.1$ & $13.4{\pm}1.8$ & $12.1{\pm}2.5$ & $11.8{\pm}2.6$ & $12.1{\pm}1.4$ \\
Kimi-K3$^{\dagger}$ ($>$2026-03-22) [$n$=236]
  & $7.2{\pm}1.2$ & $11.4{\pm}2.4$ & $9.4{\pm}2.1$ & $7.8{\pm}1.5$ & $12.0{\pm}1.6$ & $10.5{\pm}3.2$ & $9.7{\pm}1.7$ \\
GLM-5.2$^{\dagger}$ (all)
  & $8.1{\pm}0.3$ & $11.7{\pm}2.1$ & $12.3{\pm}2.2$ & $11.2{\pm}1.4$ & $11.9{\pm}0.9$ & $12.2{\pm}1.9$ & $11.2{\pm}1.5$ \\
GLM-5.2$^{\dagger}$ ($\le$2026-02-20) [$n$=321]
  & $7.0{\pm}0.2$ & $13.5{\pm}2.0$ & $17.8{\pm}2.7$ & $12.8{\pm}1.6$ & $13.0{\pm}0.6$ & $14.6{\pm}2.2$ & $13.1{\pm}3.2$ \\
GLM-5.2$^{\dagger}$ ($>$2026-02-20) [$n$=322]
  & $9.5{\pm}0.7$ & $10.1{\pm}2.4$ & $10.1{\pm}2.1$ & $9.9{\pm}1.4$ & $10.4{\pm}1.4$ & $8.0{\pm}1.4$ & $9.7{\pm}0.8$ \\
\midrule
\multicolumn{8}{@{}l@{}}{\scriptsize\textit{Multi-agent; paper side via max origin-proposer KC}} \\
multi-agent (top 4) (all)
  & $22.9{\pm}6.0$ & $36.5{\pm}11.4$ & $38.4{\pm}8.5$ & $40.1{\pm}10.3$ & $41.6{\pm}10.3$ & $36.4{\pm}10.4$ & $36.0{\pm}6.1$ \\
multi-agent (top 4) ($\le$max-origin-KC) [$n$=406]
  & $24.3{\pm}6.4$ & $38.9{\pm}12.1$ & $46.2{\pm}10.0$ & $45.4{\pm}11.0$ & $41.6{\pm}10.1$ & $39.8{\pm}10.2$ & $39.4{\pm}7.2$ \\
multi-agent (top 4) ($>$max-origin-KC) [$n$=237]
  & $19.7{\pm}5.0$ & $33.0{\pm}10.6$ & $32.9{\pm}7.5$ & $31.8{\pm}9.7$ & $41.6{\pm}11.3$ & $28.7{\pm}10.9$ & $31.3{\pm}6.5$ \\
\midrule
vs best single (all)
  & $2.5\times$ & $2.4\times$ & $2.3\times$ & $2.6\times$ & $2.6\times$ & $2.3\times$ & $2.4\times$ \\
vs best single ($\le$KC)
  & $2.4\times$ & $2.3\times$ & $2.0\times$ & $2.3\times$ & $2.5\times$ & $2.2\times$ & $2.3\times$ \\
vs best single ($>$KC)
  & $2.1\times$ & $2.3\times$ & $2.2\times$ & $2.2\times$ & $2.7\times$ & $2.2\times$ & $2.3\times$ \\
\end{longtable}
}

\newpage
{\setlength{\LTcapwidth}{\textwidth}%
\begin{longtable}{@{}>{\scriptsize}l*{7}{>{\scriptsize}c}@{}}
\caption{Per-domain $\Delta = \mathrm{Match}(\le\mathrm{KC})-\mathrm{Match}(>\mathrm{KC})$ (\%) under each model's own KC (Table~\ref{tab:contam-kc-split}).
For multi-agent, sides use the max origin-proposer KC; --- marks empty strata ($n{=}0$).
\emph{Caution:} as in Table~\ref{tab:contam-delta-A} the two sides are disjoint sets of unequal size, and here both their sizes and their membership change from row to row because each proposer has its own KC (from $n_{\le}{=}33$ for Gemini to $n_{\le}{=}407$ for Kimi; see Table~\ref{tab:contam-kc-split}).
Rows are therefore not computed on a common split and $\Delta$ values should not be compared across models as if they were; small-$n_{\le}$ rows (Gemini, DeepSeek) are especially unstable.}
\label{tab:contam-delta-B}\\
\toprule
\makebox[\deltanamew][l]{Model} & \makebox[\deltacolw]{ML} & \makebox[\deltacolw]{Astro} & \makebox[\deltacolw]{Chem} & \makebox[\deltacolw]{Mat} & \makebox[\deltacolw]{Med} & \makebox[\deltacolw]{Phys} & \makebox[\deltacolw]{Avg} \\
\midrule
\endfirsthead
\multicolumn{8}{@{}l@{}}{\scriptsize\textit{Table~\ref{tab:contam-delta-B} (continued).}}\\
\toprule
\makebox[\deltanamew][l]{Model} & \makebox[\deltacolw]{ML} & \makebox[\deltacolw]{Astro} & \makebox[\deltacolw]{Chem} & \makebox[\deltacolw]{Mat} & \makebox[\deltacolw]{Med} & \makebox[\deltacolw]{Phys} & \makebox[\deltacolw]{Avg} \\
\midrule
\endhead
\bottomrule
\endfoot
\bottomrule
\endlastfoot
Claude-Opus-4.8 & $+0.2$ & $+1.2$ & $+5.4$ & $+7.2$ & $+1.8$ & $+4.6$ & $+3.4$ \\
GPT-5.6-Sol-Pro & $+0.7$ & $+5.1$ & $+8.8$ & $+1.7$ & $+0.1$ & $+5.2$ & $+3.6$ \\
Kimi-K3 & $+2.8$ & $+0.6$ & $+3.7$ & $+4.1$ & $+0.1$ & $+1.8$ & $+2.2$ \\
GLM-5.2 & $-1.3$ & $+2.4$ & $+6.5$ & $+2.9$ & $+3.2$ & $+4.7$ & $+3.1$ \\
Gemini 3.1-Pro-Preview & $-6.0$ & $+9.0$ & --- & $+1.2$ & --- & $+1.3$ & $+0.6$ \\
DeepSeek-V4-Pro & $-0.3$ & $+4.6$ & $+4.7$ & $+2.1$ & $-7.0$ & $+1.3$ & $+0.9$ \\
Qwen3.7-Max & $-3.0$ & $+2.0$ & $+3.9$ & $+3.5$ & $+0.6$ & $+3.6$ & $+1.8$ \\
\midrule
\multicolumn{8}{@{}l@{}}{\scriptsize\textit{Top~4 proposers, judged only by the other top~3 models}} \\
Claude-Opus-4.8$^{\dagger}$ & $+0.6$ & $+2.5$ & $+4.8$ & $+6.7$ & $+2.9$ & $+4.7$ & $+3.7$ \\
GPT-5.6-Sol-Pro$^{\dagger}$ & $+0.4$ & $+4.8$ & $+8.3$ & $+1.7$ & $+0.5$ & $+5.4$ & $+3.5$ \\
Kimi-K3$^{\dagger}$ & $+2.9$ & $-0.4$ & $+4.6$ & $+5.6$ & $+0.1$ & $+1.3$ & $+2.3$ \\
GLM-5.2$^{\dagger}$ & $-2.6$ & $+3.4$ & $+7.7$ & $+2.9$ & $+2.6$ & $+6.6$ & $+3.4$ \\
\midrule
\rowcolor{magenta!8}
multi-agent (top 4) & $+4.6$ & $+6.0$ & $+13.3$ & $+13.6$ & $0.0$ & $+11.1$ & $+8.1$ \\
\end{longtable}
}

\paragraph{Takeaway.}
KC bounds put $5.1\%$--$63.3\%$ of seeds on the possibly-reachable side.
Match rates are higher on $\le$ strata on average under the fixed 2026-03-22 cut (Default mean $+2.7\%$; multi-agent $+8.5\%$; Table~\ref{tab:contam-delta-A}), and dagger agrees with Default, but the gap is domain-dependent and compares disjoint paper sets of unequal size---not a within-paper memorization effect.
The headline pattern survives the cleaner ${>}\mbox{2026-03-22}$ slice: multi-agent Match rates $19.7$--$41.6\%$ (mean $31.1\%$), and the six-domain average of vs-best-single remains $2.4\times$ (Section~\ref{sec:contam}).
Date $\le$ KC is corpus reachability, not demonstrated \mbox{instance-level} storage~\cite{morris2025memorize}; nonzero Match on the $>$ slice is the tighter anti-memorization observation.

\end{document}